\documentclass[sigconf]{acmart}

\usepackage{enumitem}
\usepackage{amsmath}
\usepackage{array}
\usepackage{multicol}
\usepackage{multirow}
\usepackage{rotating}
\usepackage{makecell}
\usepackage{colortbl}
\usepackage{xcolor}
\usepackage{algorithm}
\usepackage{algpseudocode}
\usepackage{graphicx}
\usepackage{subcaption}

\AtBeginDocument{%
  }

\copyrightyear{2025}
\acmYear{2025}
\setcopyright{acmlicensed}
\acmConference[KDD '25] {Proceedings of the 31st ACM SIGKDD Conference on Knowledge Discovery and Data Mining V.1}{August 3--7, 2025}{Toronto, ON, Canada.}
\acmBooktitle{Proceedings of the 31st ACM SIGKDD Conference on Knowledge Discovery and Data Mining V.1 (KDD '25), August 3--7, 2025, Toronto, ON, Canada}
\acmDOI{10.1145/3690624.3709252}
\acmISBN{979-8-4007-1245-6/25/08}

\begin{document}
\title{RELIEF: Reinforcement Learning Empowered Graph Feature Prompt Tuning}

\author{Jiapeng Zhu}
\affiliation{
  \institution{East China Normal University}
  \city{Shanghai}
  \country{China}}
\email{jiapengzhu@stu.ecnu.edu.cn}

\author{Zichen Ding}
\affiliation{
  \institution{East China Normal University}
  \city{Shanghai}
  \country{China}}
\email{zichending@stu.ecnu.edu.cn}

\author{Jianxiang Yu}
\affiliation{
  \institution{East China Normal University}
  \city{Shanghai}
  \country{China}}
\email{jianxiangyu@stu.ecnu.edu.cn}

\author{Jiaqi Tan}
\affiliation{
  \institution{East China Normal University}
  \city{Shanghai}
  \country{China}}
\email{jiaqitan@stu.ecnu.edu.cn}

\author{Xiang Li}
\authornote{Corresponding author}
\affiliation{
  \institution{East China Normal University}
  \institution{Engineering Research Center of Blockchain Data Management (East China Normal University), Ministry of Education}
  \institution{Shanghai Engineering Research Center of Big Data Management}
  \city{Shanghai}
  \country{China}
}
\email{xiangli@dase.ecnu.edu.cn}

\author{Weining Qian}
\affiliation{%
  \institution{East China Normal University}
  \city{Shanghai}
  \country{China}}
\email{wnqian@dase.ecnu.edu.cn}

\renewcommand{\shortauthors}{Jiapeng Zhu et al.}

\begin{abstract}
The advent of the ``pre-train, prompt'' paradigm has recently extended its generalization ability and data efficiency to graph representation learning, following its achievements in Natural Language Processing (NLP). Initial graph prompt tuning approaches tailored specialized prompting functions for Graph Neural Network (GNN) models pre-trained with specific strategies, such as edge prediction, thus limiting their applicability. In contrast, another pioneering line of research has explored universal prompting via adding prompts to the input graph's feature space, thereby removing the reliance on specific pre-training strategies. However, the necessity to add feature prompts to all nodes remains an open question. Motivated by findings from prompt tuning research in the NLP domain, which suggest that highly capable pre-trained models need less conditioning signal to achieve desired behaviors, we advocate for strategically incorporating necessary and lightweight feature prompts to certain graph nodes to enhance downstream task performance. This introduces a combinatorial optimization problem, requiring a policy to decide 1) which nodes to prompt and 2) what specific feature prompts to attach. We then address the problem by framing the prompt incorporation process as a sequential decision-making problem and propose our method, RELIEF, which employs Reinforcement Learning (RL) to optimize it. At each step, the RL agent selects a node (discrete action) and determines the prompt content (continuous action), aiming to maximize cumulative performance gain. Extensive experiments on graph and node-level tasks with various pre-training strategies in few-shot scenarios demonstrate that our RELIEF outperforms fine-tuning and other prompt-based approaches in classification performance and data efficiency. The code is available at \url{https://github.com/JasonZhujp/RELIEF}.
\end{abstract}




\maketitle

\section{Introduction}
\begin{figure*}[!t]
  \centering
  \includegraphics[width=0.95\textwidth]{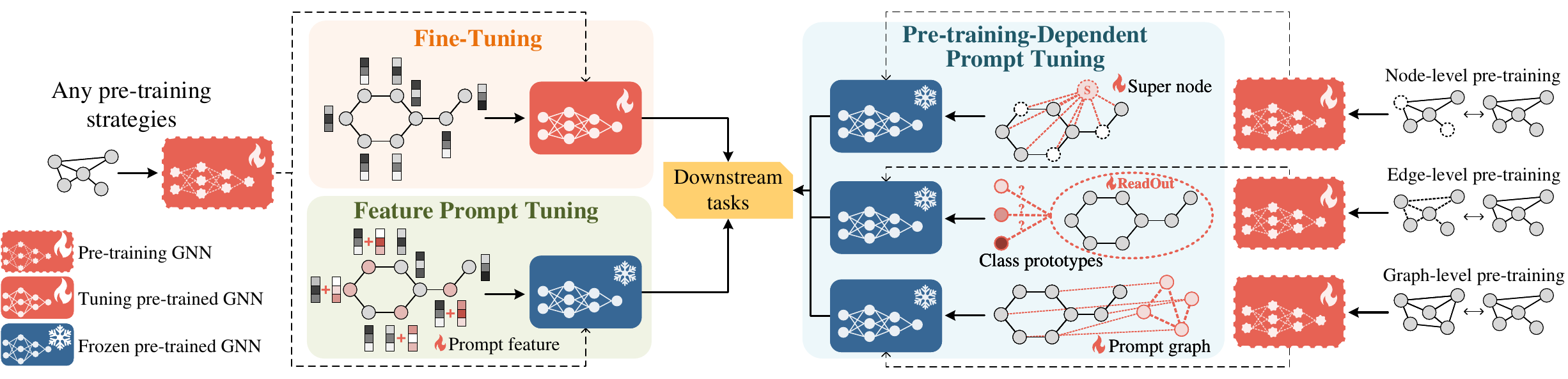}
  \vspace{-3mm}
  \caption{A Comparison of Tuning Methods. Fine-tuning (\textcolor[HTML]{E9700D}{upper left}) updates the parameters of the pre-trained GNN model. Pre-training-dependent prompt tuning (\textcolor[HTML]{205867}{right}) freezes the GNN model and requires designing specialized prompt templates aligned with pre-training strategies, whereas feature prompt tuning (\textcolor[HTML]{50632A}{lower left}) is applicable to any pre-training strategy.}
  \label{fig:tuning_comparison}
  \Description{Figure~\ref{fig:tuning_comparison} Fully described in the text.}
  \vspace{-3mm}
\end{figure*}

In recent years, Graph Neural Networks (GNNs) have been applied in diverse domains, including knowledge graphs~\cite{knowledge_graph}, social networks~\cite{social_network}, and recommender systems~\cite{recommender_system}, due to their powerful expressivity in capturing complex relationships in real-world data.
To facilitate robust graph learning under few-shot and out-of-distribution settings, significant efforts have focused on pre-training GNN models~\cite{pre-train2}.
Despite advancements, the ``pre-train, fine-tune'' paradigm still exhibits drawbacks. The misalignment between pre-training and downstream objectives can lead to sub-optimal performance~\cite{negative_transfer}. Moreover, pre-trained models are prone to overfitting given rather limited samples, leading to catastrophic forgetting~\cite{graph_catastrophic_forgetting}, which compromises their generalization ability.

To tackle the issues, prompt learning, which has shown remarkable success in Natural Language Processing (NLP)~\cite{nlp_prompt,pet1} and Computer Vision (CV)~\cite{cv_prompt1,cv_prompt2}, emerges as a promising approach. By designing informative prompts to manipulate input data, prompt tuning aligns objectives between pretext and downstream tasks, thereby augmenting the potential of pre-trained models.

Naturally, researchers have extended the ``pre-train, prompt'' framework to graph domain~\cite{graph_prompt_survey, protein, ddi, urban, ProG}. Existing methods are broadly divided into two categories based on dependence on pre-training strategies. For methods reliant on pre-training strategies, GPPT~\cite{GPPT} and GraphPrompt~\cite{Gprompt} unify pretext and downstream tasks as edge prediction. All in One~\cite{All_in_one} introduces a prompt graph to match the graph-level contrastive learning used in pre-training.
However, these methods may fail if the pre-training and prompt tuning objectives of the GNN model are misaligned.
In contrast, prompting methods agnostic to pre-training strategies offer greater compatibility. GPF~\cite{GPF} introduces graph feature-based prompt tuning, demonstrating that adding a learnable uniform feature prompt vector to each node is theoretically equivalent to any graph manipulation, without restricting pre-training strategies. GPF-plus~\cite{GPF} and SUPT~\cite{SUPT} further provide more flexible adaptation by inserting node or subgraph-specific prompted features, respectively, capturing intricate topological nuances of graph, leading to a decent performance improvement. Thus, feature-based prompting forms the basis of our paper due to its generality and effectiveness.

Nonetheless, it is critical to delve deeper into existing feature-based prompting methods. In NLP, soft prompt tuning adapts frozen language models for specific tasks using continuous prompt tokens~\cite{nlp_prompt}. Research shows that while increasing prompt token length involves more tunable parameters and thus enhances expressivity, excessively long prefixes can degrade test performance~\cite{prefix,pet2}. In fact, T5-XXL can achieve strong results even with a single-token prompt, indicating that more capable models need less conditioning signal to exhibit desired behavior~\cite{pet1}.
This finding is even more enlightening in the context of graphs. Since most GNNs follow the message-passing mechanism, information from a feature prompt attached to a target node will propagate to its neighbors, diffusing the influence of prompting signals to a broader range.

We propose that given a pre-trained GNN model, judiciously inserting minimal prompts as conditioning signals may suffice for downstream tasks. In the realm of feature prompting, ``minimal'' can be translated as an appropriate amount and magnitude of prompts.
Conversely, since prompts are directly added to node features, incorporating large-magnitude prompts to an excessive number of nodes might overwhelm the original input's feature space, causing pre-trained GNNs to perceive prompted graphs as unfamiliar, leading to a decline in transfer performance. Hence, the key lies in \textbf{strategically incorporating necessary and lightweight prompts into the original graph}, allowing a GNN model to fully leverage its learned knowledge during pre-training without being overly disrupted, and to better generalize to downstream tasks through flexible feature prompts.

The choice of prompting on certain nodes and attaching specific prompt contents leads to a combinatorial optimization problem. To tackle this, we turn to Reinforcement Learning (RL), which effectively searches for heuristic strategies by training an agent~\cite{rl_combinatorial}, and propose a \underline{\textbf{RE}}inforcement \underline{\textbf{L}}earn\underline{\textbf{I}}ng \underline{\textbf{E}}mpowered graph \underline{\textbf{F}}eature prompting method, named RELIEF. We formulate the process of inserting prompts as a sequential decision-making problem. At each step, the RL agent: 1) selects which node to prompt (discrete action), and 2) determines the prompt content, i.e., specific values of the prompt vector (continuous action). Therefore, an RL algorithm capable of handling the discrete-continuous hybrid action space~\cite{H-PPO,parameterized_rl1} is employed. The prompted graph is then evaluated by the pre-trained GNN model. At the subsequent step, a new prompt generated by the RL agent is added to the previous prompted graph. Prompt addition and performance evaluation are iteratively performed to maximize the expected cumulative performance gain on the downstream task. Techniques like policy generalization~\cite{LEEP,rl_generalization1,rl_generalization2} are integrated to ensure training stability and effectiveness. 
Our contributions are summarized as follows:
\begin{itemize}[left=0pt]
\item We suggest enhancing the performance of pre-trained GNN models on downstream tasks by adding necessary and lightweight feature prompts and design metrics to quantify their impact on inputs, offering a new perspective on graph prompt tuning.
\item As far as we know, we are the first to formulate feature prompt incorporation as a sequential decision-making problem, and propose RELIEF, an RL empowered graph feature prompting method optimized through trial and error in a hybrid action space.
\item Extensive experiments across both graph and node-level tasks with various pre-training strategies in few-shot scenarios demonstrate that RELIEF achieves superior classification performance and data efficiency compared to fine-tuning and other competitive prompt-based methods.
\end{itemize}

\section{Preliminaries}
Let an undirected graph instance be $\mathcal{G} = (\mathcal{V}, \mathcal{E}, \mathbf{X}) \in \mathbb{G}$,
where $\mathcal{V} = \{v_1, v_2, \dots, v_n\}$ denotes the node set containing $n$ nodes;
$\mathcal{E} = \{(v_i, v_j) \mid v_i, v_j \in \mathcal{V}\}$ denotes the edge set;
$\mathbf{X} = \{x_1, x_2, \dots, x_n\} \in \mathbb{R} ^ {n \times D}$ denotes the node feature matrix, where $x_i \in \mathbb{R} ^ {1 \times D}$ is the feature vector of node $v_i$ with feature dimension $D$.
The adjacency matrix $\mathbf{A} \in \{0,1\} ^ {n \times n}$ is defined such that $\mathbf{A}_{ij} = 1$ if $(v_i, v_j) \in \mathcal{E}$. 

\paragraph{\textbf{Fine-tuning and Graph Prompt Tuning.}}
Given a pre-trained GNN model $f_\theta$ and a task-specific projection head $g_{\phi}$ for making predictions, parameters $\theta$ and $\phi$ are fine-tuned on a downstream training dataset $\mathcal{D} = \{(\mathcal{G}_1, y_1), \dots, (\mathcal{G}_m, y_m)\}$ to maximize the likelihood of correctly predicting label $y$ of graph $\mathcal{G}$, described as:
\begin{displaymath}
  \max_{\theta,\phi} \prod_{i=0}^{m} P_{f_\theta, \, g_\phi} (y_i \mid \mathcal{G}_i) = \max_{\theta,\phi} \prod_{i=0}^{m} g_\phi (f_\theta (\mathbf{X}, \mathbf{A})) \left[y_i\right]
\end{displaymath}
where $\left[y_i\right]$ fetches the prediction probability of label $y_i$ from $g_\phi(\cdot)$. In contrast, the pre-trained GNN model $f_\theta$ remains frozen during prompt tuning. The primary focus shifts to learning a prompting function $u_{\psi}: \mathbb{G} \to \mathbb{G}$, which transforms the original graphs $\mathcal{G}$ to prompted graphs $u_\psi (\mathcal{G})$, used as inputs to the GNN model. The projection head $g_{\phi}$ is also tuned to coordinate with the prompting function. The optimal $u_{\psi}$ and $g_{\phi}$ are obtained by:
\begin{displaymath}
    \max_{\psi,\phi} \prod_{i=0}^{m} P_{f_\theta, \, g_\phi} (y_i \mid u_\psi (\mathcal{G}_i)) = \max_{\psi,\phi} \prod_{i=0}^{m} g_\phi (f_\theta (\mathbf{X}_\psi, \mathbf{A}_\psi)) \left[y_i\right]
\end{displaymath}
where $\mathbf{X}_\psi$ and $\mathbf{A}_\psi$ are prompted feature and adjacency matrix, respectively, acquired by prompting function $u_{\psi}$. For evaluation, the downstream testing graphs are first transformed by the prompting function and then passed through the pre-trained GNN model, followed by the projection head to produce predictions. 

\paragraph{\textbf{Graph Feature Prompt Tuning.}}
This line of prompting method introduces learnable components to the feature space of the input graph. Given the node feature matrix $\mathbf{X} = \{x_1, x_2, \dots, x_n\}$ of a graph $\mathcal{G}$, learnable prompt feature vectors $p_1, p_2, \dots , p_n \in \mathbb{R} ^ {1 \times D}$ are added to node features. This can be viewed as the prompting function $u_{\psi}$, achieving a prompted feature matrix $\mathbf{X} ^ \ast$ expressed as:
\begin{displaymath}
    \mathbf{X} ^ \ast = \{x_1 + p_1, x_2 + p_2, \dots, x_n + p_n\} = \{x^\ast_1, x^\ast_2, \dots, x^\ast_n\}
\end{displaymath}
and $\mathbf{X} ^ \ast$ is then processed by the GNN model. Note that feature prompt tuning does not explicitly prompt on the adjacency matrix, thus the representation of a prompted graph is given by $f_\theta(\mathcal{G}^\ast) = f_\theta (\mathbf{X}^\ast, \mathbf{A})$.
Different feature prompting methods vary in their prompt vector designs or learning approaches.
A comparison of aforementioned tuning methods is depicted in Figure~\ref{fig:tuning_comparison}.

\paragraph{\textbf{RL with Hybrid Action Space.}} 
Most RL algorithms operate within either discrete or continuous action spaces ~\cite{rl_book}. In our scenario, however, we deal with a hybrid action space ~\cite{parameterized_rl1}, involving selecting nodes (discrete) and specifying feature prompts (continuous). Formally, discrete actions are chosen from a finite set $\mathcal{A}_d = \{a_1, a_2, \dots, a_n\}$, each corresponding to a real-valued continuous action $z \in \mathbb{R}^D$, where $D$ is the dimension of the continuous action space $\mathcal{A}_c = \{z_1, z_2, \dots, z_n\}$. H-PPO ~\cite{H-PPO} is an RL algorithm tailored for hybrid action spaces based on actor-critic architecture. It employs two parallel actor networks for discrete and continuous actions, respectively, alongside a single critic network for value estimation. H-PPO extends Proximal Policy Optimization (PPO) ~\cite{PPO} by separately optimizing the two actors with an entropy-regularized PPO surrogate objective $\mathcal{L}^{\text{PPO}}$ to maximize expected cumulative discounted reward, while the critic minimizes Mean Squared Error (MSE) loss $\mathcal{L}^\text{Critic}$ to improve value estimation accuracy. 




\section{Method}
In this section, we first introduce the formulation of a Markov Decision Process (MDP) designed for our feature prompting scenario (Section~\ref{sec:mdp}). We then detail the architecture of the policy network to instantiate the MDP (Section~\ref{sec:architecture}). Next, we present the overall framework of RELIEF, featuring alternating training (Section~\ref{sec:framework}). To mitigate policy overfitting, we integrate policy generalization techniques (Section~\ref{sec:generalization}). Additionally, we describe two metrics designed to quantify the impact of the inserted feature prompts on original graphs (Section~\ref{sec:metric}). Note that we use graph-level classification as an example throughout this section and will later extend to node tasks in the experiments section.

\subsection{Incorporating Feature Prompts as MDP}
\label{sec:mdp}

In RL domain, the environment is typically modeled by an MDP~\cite{rl_book}. With the aim of framing the process of incorporating prompts as an MDP, we present the design of the basic elements as follows.

\paragraph{\textbf{Action Space.}}
Given a graph $\mathcal{G}$ with $n$ nodes, a discrete action $a$ is to select a node $v_a$ from the node set $\{v_1, \dots, v_n\}$, and a continuous action $z \in \mathbb{R}^{1 \times D}$ is to determine a real-valued vector for node $v_a$, where $D$ is the dimension of the initial node feature. Thus, a hybrid action, i.e., feature prompt, incorporated at time step $t$ can be denoted as $(a_t, z_t) = p^{a,z}_t$. We further define the prompt matrix at step $t$ as $\mathbf{P}_t = \{p_{1,t}, \dots, p_{n,t}\} \in \mathbb{R}^{n \times D}$. $\mathbf{P}_0$ is initialized as a zero matrix, indicating no prompt is applied. Thus, the prompted feature $\mathbf{X}^\ast$ of a prompted graph $\mathcal{G}^\ast$ is obtained by $\mathbf{X}^\ast = \mathbf{X} + \mathbf{P}$.

\paragraph{\textbf{State Transition.}} 
We define the state space as node representations of the prompted graph $\mathcal{G}^\ast$ obtained from the pre-trained GNN model $f_\theta$. This allows the agent to be aware of node representations that closely tied to the GNN model, enabling precise control.
Formally, the state at time step $t$ is denoted as:
\begin{displaymath}
\begin{split}
    s_t &\coloneqq f_\theta(\mathcal{G}^\ast_{t-1}) = f_\theta (\mathbf{X}^\ast_{t-1}, \mathbf{A}) = f_\theta (\mathbf{X} + \mathbf{P}_{t-1}, \mathbf{A}) \\
    &= \{h^\ast_{1,t-1} \, , \dots, h^\ast_{n,t-1}\} \in \mathbb{R}^{n \times d}
\end{split}
\end{displaymath}
where $h^\ast_{i,t-1}$ is the representation of node $v_i$ in the prompted graph $\mathcal{G}^\ast_{t-1}$, and $d$ is the dimension of the latent space. Notably, the current state builds upon the representation of the previous step. However, as different graphs contain varying numbers of nodes, this results in state spaces of different sizes, posing challenges for batch training. To fix the size, we set a maximum number of nodes $N$ and use zero padding, i.e., append $(N-n)$ zero vectors $\mathbf{0} \in \mathbb{R}^{1 \times d}$ to the end of $h^\ast_{n,t}$, ensuring a fixed-size state representation:
\begin{equation}
\begin{split}
\label{eq:state_computation}
    s_t &\coloneqq f_\theta (\mathbf{X}^\ast_{t-1}, \mathbf{A}) \left | \right | \mathbf{0}_{(N-n) \times d} \\
    &= \{h^\ast_{1,t-1} , \dots, h^\ast_{n,t-1}, \mathbf{0}_{n+1}, \dots, \mathbf{0}_{N}\} \in \mathbb{R}^{N \times d}
\end{split}
\end{equation}
where $||$ denotes concatenation. When the agent executes action $p^{a,z}_t$ at step $t$, which inherently operates on node $v_a$'s feature prompt $p_a$, the prompted feature matrix then transits to $\mathbf{X}^\ast_t = \mathbf{X} + \mathbf{P}_t$, with $\mathbf{P}_t$ computed as:
\begin{displaymath}
    \mathbf{P}_t = \mathbf{P}_{t-1} + p^{a,z}_t = \{p_{1,t-1} \, , \dots, \, p_{a,t-1} + p^{a,z}_t, \dots, \, p_{n,t-1}\}
\end{displaymath}
Finally, the next state $s_{t+1}$ is obtained by replacing $\mathbf{X}^\ast_{t-1}$ in Eq.~\eqref{eq:state_computation} with $\mathbf{X}^\ast_t$. Practically, this involves feeding the prompted graph $\mathcal{G}^\ast_{t}$ into the pre-trained GNN model to derive new node representations, along with zero padding, to construct the subsequent state.

\paragraph{\textbf{Reward Function.}}
An ideal reward function is goal-oriented, providing guiding signals about action values during exploration. While downstream task metrics such as AUC or F1-score might seem straightforward for graph classification tasks, we require a reward to measure the quality of prompt inserted at each step for each graph. However, these metrics are infeasible to apply to a single graph. Instead, we use the loss decrease as our instant reward, since it captures the concept of performance gain somewhat and can be derived from each graph. Formally, given two adjacent steps, the reward $r(s_t,a_t,z_t,s_{t+1})$, or $r_t$ for short, is defined as:
\begin{equation}
\label{eq:instant_reward}
    r_t = \mathcal{L}_{t-1} - \mathcal{L}_t = \mathcal{L}\left(g_\phi \left(f_\theta (\mathcal{G}^\ast_{t-1})\right),y\right) - \mathcal{L}\left(g_\phi \left(f_\theta (\mathcal{G}^\ast_t)\right),y\right)
\end{equation}
where $\mathcal{L}(\cdot)$ denotes a loss function associated with the downstream task, such as cross-entropy loss for classification. Hence, the reward is positive if the prompt assigned at this step leads to a loss decrease, and negative if it increases loss. Consequently, the cumulative reward reflects the total loss decrease from time step $1$ to $T$, serving as a surrogate measurement of overall performance improvement.
\subsection{Policy Network Architecture}
\label{sec:architecture}
\begin{figure*}[!t]
  \centering
  \includegraphics[width=0.82\textwidth]{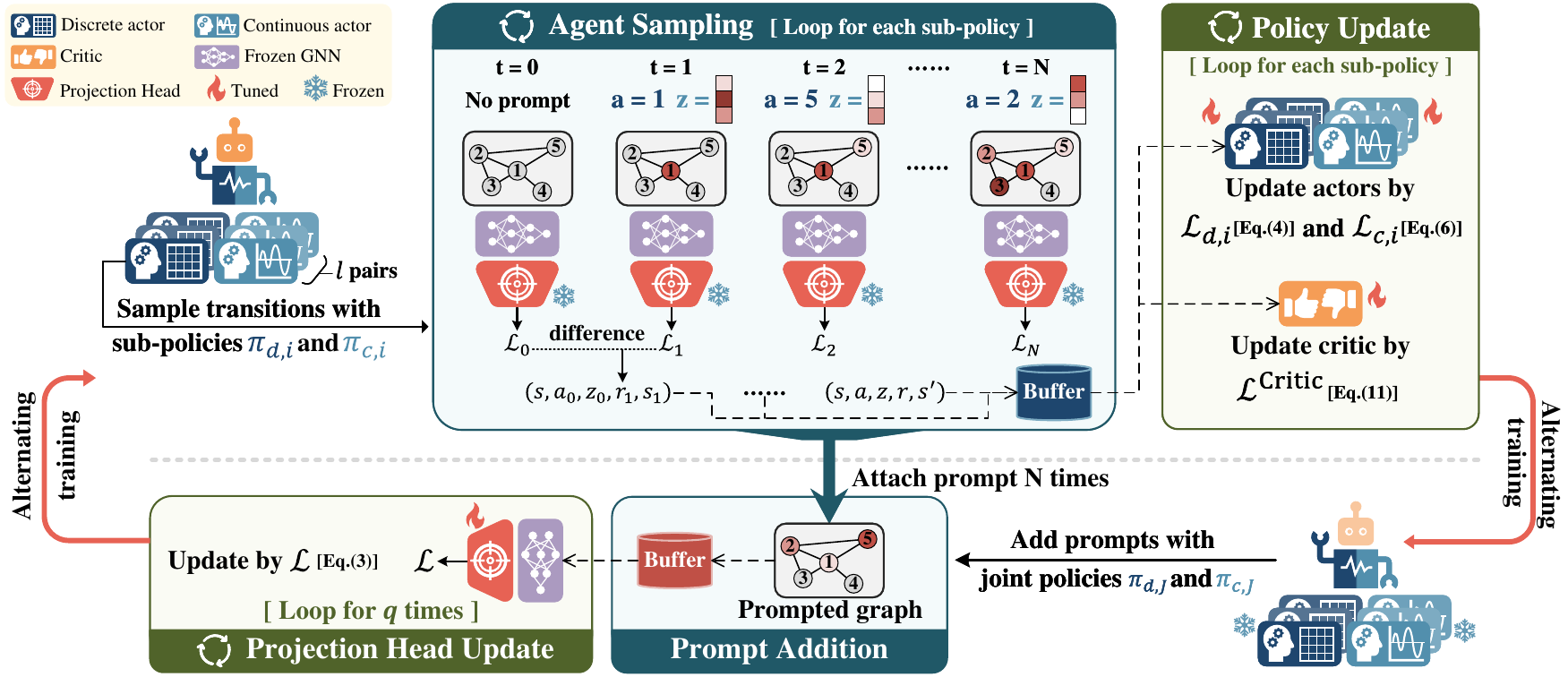}
  \vspace{-0.5em}
  \caption{RELIEF pipeline. The policy network (upper) and the projection head (lower) are trained alternately. Feature prompts are incorporated during the agent sampling and prompt addition processes via discrete and continuous actors.}
  \label{fig:pipeline}
  \Description{Figure~\ref{fig:pipeline} Fully described in the text.}
  \vspace{-1em}
\end{figure*}

RELIEF employs H-PPO, comprising two parallel actor networks and a single critic network collectively referred to as the policy network $\Pi_\omega$, where $\omega$ denotes the parameters of these networks.

In practice, all the three networks share the first few layers to encode state information. Since the state space is designed as the node representations of a prompted graph, we utilize the pre-trained GNN model $f_\theta$ as the state encoder. Thereafter, Multi-Layer Perceptrons (MLPs) with different output dimensions are connected to each of the three networks, facilitating the corresponding functions. Given a prompted graph $\mathcal{G}^\ast$, the forward propagation of the networks is presented as follows:
\begin{displaymath}
\begin{split}
    p(a \mid s) &\gets \textsc{Softmax}\left( \text{MLP}_a \left( f_\theta(\mathcal{G}^\ast) \right) \right) \\
    \boldsymbol{\mu}(s,a) &\gets \text{MLP}_z \left( f_\theta(\mathcal{G}^\ast) \right) \left[a\right] \\
    V(s) &\gets \text{MLP}_c \left( \textsc{Flatten}\left( f_\theta(\mathcal{G}^\ast) \right) \right)
\end{split}
\end{displaymath}
\vspace{-2mm}

Let's elaborate on each expression in turn.
\vspace{-2mm}

\paragraph{\textbf{Discrete Actor}}
Represents the discrete policy $\pi_d(a|s)$. Given node representations of a prompted graph as state $s$, i.e., $f_\theta(\mathcal{G}^\ast)$ with size $ N \times d$, the $\text{MLP}_a$ followed by a \textsc{Softmax} operation transforms $s$ into a discrete action probability $p(a|s) \in \mathbb{R}^n$. The agent then either samples a node $v_a$ based on this probability or greedily selects the node with the highest probability as a discrete action, indicating a stochastic or deterministic policy, respectively. 
Note that we set the entries in $p(a|s)$ associated with zero padding to zero, eliminating their chance of being selected, which reduces the number of valid discrete actions from $N$ to $n$.

\paragraph{\textbf{Continuous Actor}}
Represents the continuous policy $\pi_c(z|s,a)$. Given state $s \in \mathbb{R} ^ {N \times d}$, the $\text{MLP}_z$ outputs $N$ parameters $\boldsymbol{\mu} \in \mathbb{R} ^ {1 \times D}$ for all the $N$ possible discrete actions. 
Then $\boldsymbol{\mu}$ with index $\left[a\right]$ is chosen paired with the selected discrete action $a$. 
Subsequently, the agent constructs a Gaussian distribution based on $(\boldsymbol{\mu}, \boldsymbol{\sigma})$ and samples a vector $z \in \mathbb{R} ^ {1 \times D}$ as a prompt feature $p^{a,z}$ stochastically, or directly use $\boldsymbol{\mu}$ as the action deterministically, where the standard deviation $\boldsymbol{\sigma} \in \mathbb{R} ^ {1 \times D}$ can either be learned or pre-defined. To ensure $p^{a,z}$ remains within a desired range, each dimension of $z$ is clamped to range $\left[-z_\text{max}, z_\text{max} \right]$, where $z_\text{max} > 0$ is a hyper-parameter controlling the scale of the prompt added at every step.

\paragraph{\textbf{Critic}}
Used for estimating the state-value function. At its core, it maps a state $s$ to a real value $V(s) \in \mathbb{R}$. However, there exists a dimension discrepancy: our state space possesses a node-level granularity, whereas value estimation is based on a global view. Therefore, we use \textsc{Flatten} operation to transform the size of state $s$ from $N \times d$ to $1 \times Nd$. The flattened vector is then passed through $\text{MLP}_c$ to produce a real value, treated as the estimation of $V(s)$.

It is worth noting that the state encoder in the policy network $\Pi_\omega$ is a direct copy of the pre-trained GNN model and keeps frozen during policy learning. This implies it is by updating the parameters of the MLPs, that the actors are enabled to map states to actions, and the critic is enabled to map states to state values. Such structure has been widely used in RL from Human Feedback (RLHF) practice~\cite{Instruct-GPT, safe_rlhf}, where reward models and policy networks are constructed on fixed LLMs with learnable linear layers appended.

\subsection{Overall Framework of RELIEF}
\label{sec:framework}

RELIEF contains two trainable modules: the policy network and the projection head, as illustrated in Figure~\ref{fig:pipeline}. Effective coordination of these two modules greatly enhances the prompting performance. 
We train them alternatively in practice, as described below.

\subsubsection{Policy network training}
Given the frozen pre-trained GNN model $f_\theta$, the policy network $\Pi_\omega$, the projection head $g_\phi$ and a graph $\mathcal{G}$ with label $y$ containing $n$ nodes, the initial loss $\mathcal{L}_0$ is computed via $\mathcal{L} (g_\phi ( f_\theta (\mathcal{G}) ), y)$, with no prompt attached.

At each step, the agent samples a feature prompt $p^{a,z}_t$ with respect to the policies $\pi_d$ and $\pi_c$, and adds it to node $v_a$. The prompted graph is then passed to the GNN model followed by the projection head to obtain the current loss according to label $y$. By comparing the current loss with the previous one, an instant reward is computed by Eq.~\eqref{eq:instant_reward}. As a result, the agent collects a transition, denoted as a tuple $(s, a, z, r, s')$. The prompt addition process is repeated $n$ times, equivalent to the number of nodes, theoretically giving every node a chance of being prompted. Notably, both actors use stochastic policies for better exploration during training.

Next, the collected $n$ transitions are used to update the policy network. The two actors are independently trained using the PPO surrogate objective $\mathcal{L}^\text{PPO}$, whereas the critic is trained by MSE loss $\mathcal{L}^\text{Critic}$. The above process is batchified by consuming a batch of graphs simultaneously to increase sampling and training efficiency.

\subsubsection{Projection head training}
This aligns predictions with their correct labels by coordinating the projection head with representations of the ultimate prompted graphs. Thus, we use the learned policy to insert feature prompts for $n$ steps to obtain the final prompted graph $\mathcal{G}^\ast$.
Here, we modify the continuous policy to be deterministic, ensuring identical prompt vector values for the same state and discrete action. This guarantees stable prompted graphs and thus consistent representations, which together with the labels, are used to supervise the projection head update.
Given $m$ graph samples, the projection head $g_\phi$ is updated by minimizing the objective:
\begin{equation}
\label{eq:projection_head_objective}
    \min_{\phi}{ \frac{1}{m} \sum^{m}_{i=1} \mathcal{L}\left(g_\phi \left( f_\theta(\mathcal{G}^\ast_i) \right), y \right)}
\end{equation}
where $\mathcal{L}(\cdot)$ is the same loss function as in Eq.~\eqref{eq:instant_reward}. To facilitate a more rapid coordination of the projection head with the policy, the projection head is updated $q$ times.

To summarize, the above alternating process -- training policy network once and projection head $q$ times -- defines a training epoch. At the evaluation stage, we directly apply the two learned actors to incorporate feature prompts step by step into the downstream validation and testing graphs. These prompted graphs are transformed by the pre-trained GNN model and the learned projection head to produce predictions, which are then measured by downstream metrics.
\subsection{Policy Generalization}
\label{sec:generalization}
When trained in a limited set of environments, general RL algorithms often exhibit overfitting, resulting in poor generalization to unseen testing scenarios~\cite{rl_generalization1,rl_generalization2}, which is exactly the case in few-shot prompt tuning. To tackle this, we introduce a policy generalization technique, named LEEP~\cite{LEEP}, which can be seamlessly integrated with PPO, and thus compatible to our RELIEF.

Practically, LEEP is an ensemble-based method designed for discrete action spaces, adding a regularization term to PPO surrogate objective $\mathcal{L}^\text{PPO}$ for updating the actor network. LEEP learns a generalized joint policy by leveraging all sub-policies. To generalize our discrete policy $\pi_d(a|s)$, we learn $l$ discrete sub-policies $\{\pi_{d,1}, \dots, \pi_{d,l}\}$. Each $\pi_{d,i}$ gathers transitions from a training graph set $\mathcal{D}_i$, which is bootstrap-sampled from the entire training set $\mathcal{D}$. Each $\pi_{d,i}$ is updated by maximizing the expected reward, while minimizing the disagreement between $\pi_{d,i}$ itself and the joint discrete policy $\pi_{d,\,J}$, i.e., maximizing the following objective:
\begin{equation}
\label{eq:overall_discrete_objective}
    \mathcal{L}_{d,i} = \mathcal{L}^\text{PPO}_{d,i} - \alpha_d \mathbb{E}_{s \sim \pi_{d,i}, \, \mathcal{D}_i} \left[ D_\text{KL} \Big( \pi_{d,i}(a|s) \left| \right| \pi_{d,\,J} (a|s) \Big) \right]
\end{equation}
where $\alpha_d > 0$ is a uniform penalty hyper-parameter for all discrete sub-polices and $D_{\text{KL}}$ denotes the KL-divergence. 
The discrete joint policy $\pi_{d,\,J}$ combines all sub-policies $\pi_{d,i}$, and is computed as:
\begin{equation}
\label{eq:joint_discrete_policy}
    \pi_{d,\,J}(a|s) = \frac{\max_{i=1,\dots,l} \pi_{d,i}(a|s)}{\sum_{a'} \max_{i=1,\dots,l} \pi_{d,i}(a'|s)}
\end{equation}
indicating that to obtain the discrete action probability $p(a|s)$ given by $\pi_{d,\,J}$, we take the maximum probability across all $l$ sub-policies for each action $a$, and normalize these maxima
by their sum across all actions.

Since RELIEF requires hybrid action spaces, we extend LEEP to continuous action spaces. Similarly, we learn $l$ continuous sub-policies $\{\pi_{c,1}, \dots, \pi_{c,l}\}$, each exploring in its respective training contexts $\mathcal{D}_i$. In other words, we employ $l$ parallel H-PPO algorithms, but still with a single critic. Each continuous sub-policy $\pi_{c,i}$ is trained by maximizing the following objective:
\begin{equation}
\label{eq:overall_continuous_objective}
    \mathcal{L}_{c,i} = \mathcal{L}^\text{PPO}_{c,i} - \alpha_c \mathbb{E}_{s \sim \pi_{c,i}, \, \mathcal{D}_i} \left[ D_\text{KL} \Big( \pi_{c,i}(a|s) \left| \right| \pi_{c,\,J} (a|s) \Big) \right]
\end{equation}
where $\alpha_c > 0$ is a uniform penalty hyper-parameter for continuous sub-policies. The continuous joint policy $\pi_{c,\,J}$ is defined as:
\begin{equation}
\label{eq:joint_continuous_policy}
    \pi_{c,\,J}(z|s,a) =  \frac{1}{l} \sum_{i=1}^l \pi_{c,i}(z|s,a) = \frac{1}{l} \sum_{i=1}^l \boldsymbol{\mu}_i (s,a)
\end{equation}
indicating that the average mean $\boldsymbol{\mu}_i$ of all sub-policies is used as the continuous joint policy. 

To sum up, the policy network, enhanced by policy generalization techniques, now comprises $l$ discrete actors, $l$ continuous actors and a single critic. During training, discrete and continuous actors within each pair are independently updated using Eq.~\eqref{eq:overall_discrete_objective} and~\eqref{eq:overall_continuous_objective}, respectively, followed by the critic update using $\mathcal{L}^\text{Critic}$. Each pair of actors is sequentially updated. The joint policies $\pi_{d,\,J}$ and $\pi_{c,\,J}$ are applied to incorporate prompt features when training the projection head or testing at the evaluation stage.
\subsection{Metrics for Quantifying Prompts Impact}
\label{sec:metric}

Our RELIEF stands out from existing feature prompting methods~\cite{GPF,SUPT} by incorporating necessary and lightweight feature prompts. To measure the disturbance to the original input space caused by these prompts, we introduce two metrics: Prompt Coverage Ratio (PCR) and Average Prompt Magnitude (APM).

\paragraph{\textbf{Prompt Coverage Ratio.}}
Given a graph $\mathcal{G}$ with $n$ nodes, and the final prompt matrix $\mathbf{P} = \{p_1,\dots,p_n\} \in \mathbb{R} ^ {n \times D}$ obtained by adding feature prompts $p^{a,z}$ over $n$ steps, starting from a zero matrix, PCR is calculated as:
\begin{equation}
    \text{PCR}(\mathcal{G}) = \frac{1}{n} \sum_{i=1}^n \mathbf{1} \left[ p_i \ne \mathbf{0}_{1 \times D}\right] \; \in \left[0,1\right]
\end{equation}
where $\mathbf{1}\left[\cdot\right]$ is the indicator function, which equals 1 if the feature prompt $p_i$ is not a zero vector $\mathbf{0}_{1 \times D}$ (i.e., a valid prompt), otherwise 0. PCR signifies the proportion of nodes prompted at least once during the entire prompt addition process.

\paragraph{\textbf{Average Prompt Magnitude.}}
We use L1-norm of the valid feature prompts averaged over dimensions, to describe the magnitude of inserted prompts, which is computed as:
\begin{equation}
    \text{APM}(\mathcal{G}) = \frac{1}{n} \sum_{i=1}^n  \frac{1}{D} \mathbf{1} \left[ p_i \ne \mathbf{0}_{1 \times D}\right] \cdot \left\| p_i  \right\|_1 \; \in \left[0, +\infty \right)
\end{equation}
where $\left\| p_i \right\|_1$ denotes the Manhattan norm of $p_i$, i.e., the sum of the absolute values of all entries of $p_i$. Thus, APM characterises the scale of valid feature prompts, offering a tangible measurement in terms of ``lightweight''.

PCR and APM collectively assess how extensively and significantly feature prompts are integrated, which are applicable for evaluating any feature prompting methods.
\begin{table*}[!htbp]
  \caption{ROC-AUC (\%) and standard deviation for graph classification on molecule property prediction benchmark under 50-shot scenario with various pre-training and tuning strategies. The best results for each dataset and pre-training strategy are highlighted in bold, and the runner-up is underlined.}
  \vspace{-0.8em}
  \resizebox{\linewidth}{!}{
      \centering
      \fontsize{6.2pt}{6.2pt}\selectfont
      \begin{tabular}{ccccccccccc}
        \toprule
        & \makecell{Tuning \\ Strategy} & BBBP & Tox21 & ToxCast & SIDER & ClinTox & MUV & HIV & BACE & \textbf{Avg.} \\
    
        \midrule
    
        \multirow{6}{*}{\begin{sideways}Infomax\end{sideways}} &
        FT & 
        65.26 {\scriptsize \textcolor{gray}{$\pm$1.05}} &
        71.54 {\scriptsize \textcolor{gray}{$\pm$0.73}} &
        57.98 {\scriptsize \textcolor{gray}{$\pm$0.42}} &
        53.45 {\scriptsize \textcolor{gray}{$\pm$0.49}} &
        59.01 {\scriptsize \textcolor{gray}{$\pm$2.18}} &
        64.08 {\scriptsize \textcolor{gray}{$\pm$0.84}} &
        65.13 {\scriptsize \textcolor{gray}{$\pm$2.50}} &
        64.57 {\scriptsize \textcolor{gray}{$\pm$0.71}} & 
        62.63 \\
        & GPF &
        66.30 {\scriptsize \textcolor{gray}{$\pm$0.94}} &
        71.26 {\scriptsize \textcolor{gray}{$\pm$0.30}} &
        58.33 {\scriptsize \textcolor{gray}{$\pm$0.27}} &
        53.65 {\scriptsize \textcolor{gray}{$\pm$0.34}} &
        63.11 {\scriptsize \textcolor{gray}{$\pm$2.35}} &
        67.19 {\scriptsize \textcolor{gray}{$\pm$0.10}} &
        64.09 {\scriptsize \textcolor{gray}{$\pm$0.66}} &
        64.32 {\scriptsize \textcolor{gray}{$\pm$0.88}} & 
        63.53 \\
        & GPF-plus &
        66.12 {\scriptsize \textcolor{gray}{$\pm$1.27}} &
        71.54 {\scriptsize \textcolor{gray}{$\pm$0.63}} &
        58.43 {\scriptsize \textcolor{gray}{$\pm$0.27}} &
        53.76 {\scriptsize \textcolor{gray}{$\pm$1.15}} &
        65.06 {\scriptsize \textcolor{gray}{$\pm$1.56}} &
        \underline{67.20} {\scriptsize \textcolor{gray}{$\pm$0.17}} &
        64.81 {\scriptsize \textcolor{gray}{$\pm$1.62}} &
        65.88 {\scriptsize \textcolor{gray}{$\pm$1.72}} & 
        64.10 \\
        & SUPT\textsubscript{soft} &
        \underline{66.45} {\scriptsize \textcolor{gray}{$\pm$0.85}} &
        \textbf{71.82} {\scriptsize \textcolor{gray}{$\pm$0.16}} &
        \underline{58.71} {\scriptsize \textcolor{gray}{$\pm$0.44}} &
        53.72 {\scriptsize \textcolor{gray}{$\pm$0.45}} &
        \underline{65.96} {\scriptsize \textcolor{gray}{$\pm$3.37}} &
        66.52 {\scriptsize \textcolor{gray}{$\pm$0.55}} &
        65.51 {\scriptsize \textcolor{gray}{$\pm$2.31}} &
        66.26 {\scriptsize \textcolor{gray}{$\pm$2.45}} & 
        \underline{64.37} \\
        & SUPT\textsubscript{hard} &
        66.14 {\scriptsize \textcolor{gray}{$\pm$0.85}} &
        71.55 {\scriptsize \textcolor{gray}{$\pm$0.37}} &
        58.65 {\scriptsize \textcolor{gray}{$\pm$0.33}} &
        \underline{53.82} {\scriptsize \textcolor{gray}{$\pm$0.49}} &
        65.90 {\scriptsize \textcolor{gray}{$\pm$3.41}} &
        66.53 {\scriptsize \textcolor{gray}{$\pm$0.35}} &
        \underline{65.59} {\scriptsize \textcolor{gray}{$\pm$2.26}} &
        \underline{66.40} {\scriptsize \textcolor{gray}{$\pm$0.70}} & 
        64.32 \\
        & \cellcolor{gray!30}{RELIEF} &
        \textbf{67.93} {\scriptsize \textcolor{gray}{$\pm$0.73}} &
        \underline{71.58} {\scriptsize \textcolor{gray}{$\pm$0.13}} &
        \textbf{58.78} {\scriptsize \textcolor{gray}{$\pm$0.17}} &
        \textbf{53.95} {\scriptsize \textcolor{gray}{$\pm$0.40}} &
        \textbf{66.28} {\scriptsize \textcolor{gray}{$\pm$2.07}} &
        \textbf{67.23} {\scriptsize \textcolor{gray}{$\pm$0.08}} &
        \textbf{67.00} {\scriptsize \textcolor{gray}{$\pm$0.55}} &
        \textbf{67.95} {\scriptsize \textcolor{gray}{$\pm$0.52}} & 
        \textbf{65.09} \\
    
        \addlinespace[1pt]
        \midrule[0.1pt]
        \addlinespace[2pt]
    
        \multirow{6}{*}{\begin{sideways}AttrMasking\end{sideways}} &
        FT & 
        66.48 {\scriptsize \textcolor{gray}{$\pm$0.44}} &
        72.32 {\scriptsize \textcolor{gray}{$\pm$0.19}} &
        57.35 {\scriptsize \textcolor{gray}{$\pm$0.42}} &
        54.62 {\scriptsize \textcolor{gray}{$\pm$0.58}} &
        59.71 {\scriptsize \textcolor{gray}{$\pm$4.43}} &
        61.49 {\scriptsize \textcolor{gray}{$\pm$1.04}} &
        58.37 {\scriptsize \textcolor{gray}{$\pm$1.51}} &
        65.41 {\scriptsize \textcolor{gray}{$\pm$2.09}} & 
        61.97 \\
        & GPF &
        66.67 {\scriptsize \textcolor{gray}{$\pm$0.60}} &
        72.31 {\scriptsize \textcolor{gray}{$\pm$0.30}} &
        58.01 {\scriptsize \textcolor{gray}{$\pm$0.27}} &
        \underline{55.65} {\scriptsize \textcolor{gray}{$\pm$1.92}} &
        67.25 {\scriptsize \textcolor{gray}{$\pm$2.41}} &
        62.58 {\scriptsize \textcolor{gray}{$\pm$0.31}} &
        58.71 {\scriptsize \textcolor{gray}{$\pm$2.56}} &
        68.54 {\scriptsize \textcolor{gray}{$\pm$1.24}} & 
        63.72 \\
        & GPF-plus &
        66.29 {\scriptsize \textcolor{gray}{$\pm$0.36}} &
        \underline{72.75} {\scriptsize \textcolor{gray}{$\pm$0.38}} &
        57.91 {\scriptsize \textcolor{gray}{$\pm$0.38}} &
        55.05 {\scriptsize \textcolor{gray}{$\pm$1.20}} &
        \underline{68.89} {\scriptsize \textcolor{gray}{$\pm$2.86}} &
        62.70 {\scriptsize \textcolor{gray}{$\pm$0.48}} &
        58.71 {\scriptsize \textcolor{gray}{$\pm$1.07}} &
        68.07 {\scriptsize \textcolor{gray}{$\pm$1.20}} & 
        63.80 \\
        & SUPT\textsubscript{soft} &
        66.28 {\scriptsize \textcolor{gray}{$\pm$0.81}} &
        \textbf{72.76} {\scriptsize \textcolor{gray}{$\pm$0.41}} &
        \underline{58.28} {\scriptsize \textcolor{gray}{$\pm$0.34}} &
        54.70 {\scriptsize \textcolor{gray}{$\pm$1.03}} &
        68.58 {\scriptsize \textcolor{gray}{$\pm$4.54}} &
        \underline{62.81} {\scriptsize \textcolor{gray}{$\pm$0.25}} &
        58.95 {\scriptsize \textcolor{gray}{$\pm$1.71}} &
        66.41 {\scriptsize \textcolor{gray}{$\pm$0.49}} & 
        63.60 \\
        & SUPT\textsubscript{hard} &
        \underline{66.93} {\scriptsize \textcolor{gray}{$\pm$0.91}} &
        \underline{72.75} {\scriptsize \textcolor{gray}{$\pm$0.41}} &
        58.18 {\scriptsize \textcolor{gray}{$\pm$0.44}} &
        55.20 {\scriptsize \textcolor{gray}{$\pm$1.26}} &
        68.36 {\scriptsize \textcolor{gray}{$\pm$3.57}} &
        \underline{62.81} {\scriptsize \textcolor{gray}{$\pm$0.22}} &
        \underline{59.02} {\scriptsize \textcolor{gray}{$\pm$1.61}} &
        \underline{68.92} {\scriptsize \textcolor{gray}{$\pm$0.96}} & 
        \underline{64.02} \\
        & \cellcolor{gray!30}{RELIEF} &
        \textbf{67.18} {\scriptsize \textcolor{gray}{$\pm$0.55}} &
        72.40 {\scriptsize \textcolor{gray}{$\pm$0.32}} &
        \textbf{58.41} {\scriptsize \textcolor{gray}{$\pm$0.14}} &
        \textbf{56.54} {\scriptsize \textcolor{gray}{$\pm$0.75}} &
        \textbf{74.39} {\scriptsize \textcolor{gray}{$\pm$1.18}} &
        \textbf{63.18} {\scriptsize \textcolor{gray}{$\pm$0.17}} &
        \textbf{59.42} {\scriptsize \textcolor{gray}{$\pm$0.55}} &
        \textbf{69.37} {\scriptsize \textcolor{gray}{$\pm$0.70}} & 
        \textbf{65.11} \\
    
        \addlinespace[1pt]
        \midrule[0.1pt]
        \addlinespace[2pt]
    
        \multirow{6}{*}{\begin{sideways}ContextPred\end{sideways}} &
        FT & 
        \underline{62.82} {\scriptsize \textcolor{gray}{$\pm$0.83}} &
        70.11 {\scriptsize \textcolor{gray}{$\pm$0.38}} &
        57.68 {\scriptsize \textcolor{gray}{$\pm$0.69}} &
        56.68 {\scriptsize \textcolor{gray}{$\pm$0.78}} &
        \underline{60.99} {\scriptsize \textcolor{gray}{$\pm$2.28}} &
        61.28 {\scriptsize \textcolor{gray}{$\pm$1.93}} &
        59.14 {\scriptsize \textcolor{gray}{$\pm$1.93}} &
        64.47 {\scriptsize \textcolor{gray}{$\pm$2.47}} & 
        61.65 \\
        & GPF &
        61.65 {\scriptsize \textcolor{gray}{$\pm$0.76}} &
        \underline{70.42} {\scriptsize \textcolor{gray}{$\pm$0.29}} &
        \underline{58.51} {\scriptsize \textcolor{gray}{$\pm$0.38}} &
        56.55 {\scriptsize \textcolor{gray}{$\pm$0.46}} &
        58.80 {\scriptsize \textcolor{gray}{$\pm$2.57}} &
        \underline{64.83} {\scriptsize \textcolor{gray}{$\pm$0.32}} &
        \underline{59.68} {\scriptsize \textcolor{gray}{$\pm$2.20}} &
        68.57 {\scriptsize \textcolor{gray}{$\pm$1.19}} & 
        62.38 \\
        & GPF-plus &
        61.25 {\scriptsize \textcolor{gray}{$\pm$0.73}} &
        70.12 {\scriptsize \textcolor{gray}{$\pm$0.42}} &
        57.64 {\scriptsize \textcolor{gray}{$\pm$0.70}} &
        \underline{56.86} {\scriptsize \textcolor{gray}{$\pm$0.52}} &
        61.43 {\scriptsize \textcolor{gray}{$\pm$3.06}} &
        64.59 {\scriptsize \textcolor{gray}{$\pm$0.28}} &
        59.15 {\scriptsize \textcolor{gray}{$\pm$2.23}} &
        68.58 {\scriptsize \textcolor{gray}{$\pm$1.42}} & 
        \underline{62.45} \\
        & SUPT\textsubscript{soft} &
        61.85 {\scriptsize \textcolor{gray}{$\pm$1.56}} &
        70.37 {\scriptsize \textcolor{gray}{$\pm$0.14}} &
        57.82 {\scriptsize \textcolor{gray}{$\pm$0.37}} &
        56.57 {\scriptsize \textcolor{gray}{$\pm$0.50}} &
        59.13 {\scriptsize \textcolor{gray}{$\pm$1.99}} &
        64.59 {\scriptsize \textcolor{gray}{$\pm$1.00}} &
        56.58 {\scriptsize \textcolor{gray}{$\pm$1.94}} &
        68.92 {\scriptsize \textcolor{gray}{$\pm$0.92}} & 
        61.98 \\
        & SUPT\textsubscript{hard} &
        62.29 {\scriptsize \textcolor{gray}{$\pm$1.41}} &
        70.40 {\scriptsize \textcolor{gray}{$\pm$0.15}} &
        58.06 {\scriptsize \textcolor{gray}{$\pm$0.39}} &
        56.34 {\scriptsize \textcolor{gray}{$\pm$0.72}} &
        59.18 {\scriptsize \textcolor{gray}{$\pm$2.07}} &
        64.57 {\scriptsize \textcolor{gray}{$\pm$0.69}} &
        59.08 {\scriptsize \textcolor{gray}{$\pm$1.47}} &
        \underline{69.40} {\scriptsize \textcolor{gray}{$\pm$1.62}} & 
        62.42 \\
        & \cellcolor{gray!30}{RELIEF} &
        \textbf{62.99} {\scriptsize \textcolor{gray}{$\pm$0.67}} &
        \textbf{70.51} {\scriptsize \textcolor{gray}{$\pm$0.14}} &
        \textbf{58.58} {\scriptsize \textcolor{gray}{$\pm$0.04}} &
        \textbf{56.89} {\scriptsize \textcolor{gray}{$\pm$0.25}} &
        \textbf{63.06} {\scriptsize \textcolor{gray}{$\pm$1.74}} &
        \textbf{64.87} {\scriptsize \textcolor{gray}{$\pm$0.31}} &
        \textbf{59.72} {\scriptsize \textcolor{gray}{$\pm$1.67}} &
        \textbf{70.19} {\scriptsize \textcolor{gray}{$\pm$0.96}} & 
        \textbf{63.35} \\
    
        \addlinespace[1pt]
        \midrule[0.1pt]
        \addlinespace[2pt]
    
        \multirow{6}{*}{\begin{sideways}GCL\end{sideways}} &
        FT & 
        62.13 {\scriptsize \textcolor{gray}{$\pm$1.66}} &
        61.35 {\scriptsize \textcolor{gray}{$\pm$0.88}} &
        53.96 {\scriptsize \textcolor{gray}{$\pm$0.80}} &
        52.63 {\scriptsize \textcolor{gray}{$\pm$0.71}} &
        68.73 {\scriptsize \textcolor{gray}{$\pm$4.74}} &
        51.96 {\scriptsize \textcolor{gray}{$\pm$1.93}} &
        58.67 {\scriptsize \textcolor{gray}{$\pm$1.41}} &
        49.94 {\scriptsize \textcolor{gray}{$\pm$3.80}} & 
        57.42 \\
        & GPF &
        61.58 {\scriptsize \textcolor{gray}{$\pm$1.81}} &
        59.92 {\scriptsize \textcolor{gray}{$\pm$1.29}} &
        54.44 {\scriptsize \textcolor{gray}{$\pm$0.31}} &
        51.21 {\scriptsize \textcolor{gray}{$\pm$0.56}} &
        75.52 {\scriptsize \textcolor{gray}{$\pm$3.08}} &
        52.26 {\scriptsize \textcolor{gray}{$\pm$1.28}} &
        58.37 {\scriptsize \textcolor{gray}{$\pm$1.16}} &
        56.09 {\scriptsize \textcolor{gray}{$\pm$2.44}} & 
        58.67 \\
        & GPF-plus &
        62.19 {\scriptsize \textcolor{gray}{$\pm$1.45}} &
        60.13 {\scriptsize \textcolor{gray}{$\pm$0.52}} &
        54.43 {\scriptsize \textcolor{gray}{$\pm$0.43}} &
        50.90 {\scriptsize \textcolor{gray}{$\pm$0.78}} &
        75.50 {\scriptsize \textcolor{gray}{$\pm$1.11}} &
        52.40 {\scriptsize \textcolor{gray}{$\pm$1.55}} &
        58.30 {\scriptsize \textcolor{gray}{$\pm$1.68}} &
        59.55 {\scriptsize \textcolor{gray}{$\pm$3.17}} & 
        59.18 \\
        & SUPT\textsubscript{soft} &
        63.96 {\scriptsize \textcolor{gray}{$\pm$0.85}} &
        60.13 {\scriptsize \textcolor{gray}{$\pm$0.52}} &
        54.57 {\scriptsize \textcolor{gray}{$\pm$0.76}} &
        51.44 {\scriptsize \textcolor{gray}{$\pm$1.24}} &
        \underline{78.03} {\scriptsize \textcolor{gray}{$\pm$2.32}} &
        51.86 {\scriptsize \textcolor{gray}{$\pm$1.48}} &
        59.30 {\scriptsize \textcolor{gray}{$\pm$2.59}} &
        57.39 {\scriptsize \textcolor{gray}{$\pm$2.64}} & 
        59.59 \\
        & SUPT\textsubscript{hard} &
        \underline{64.15} {\scriptsize \textcolor{gray}{$\pm$0.96}} &
        60.56 {\scriptsize \textcolor{gray}{$\pm$0.20}} &
        \underline{54.72} {\scriptsize \textcolor{gray}{$\pm$0.71}} &
        51.60 {\scriptsize \textcolor{gray}{$\pm$0.85}} &
        77.88 {\scriptsize \textcolor{gray}{$\pm$2.06}} &
        51.78 {\scriptsize \textcolor{gray}{$\pm$1.72}} &
        59.24 {\scriptsize \textcolor{gray}{$\pm$2.42}} &
        \underline{59.68} {\scriptsize \textcolor{gray}{$\pm$1.46}} & 
        \underline{59.95} \\
        & All in One &
        62.90 {\scriptsize \textcolor{gray}{$\pm$3.77}} &
        \textbf{61.49} {\scriptsize \textcolor{gray}{$\pm$0.96}} &
        \underline{54.72} {\scriptsize \textcolor{gray}{$\pm$1.03}} &
        \underline{52.73} {\scriptsize \textcolor{gray}{$\pm$0.84}} &
        66.59 {\scriptsize \textcolor{gray}{$\pm$3.84}} &
        \underline{53.34} {\scriptsize \textcolor{gray}{$\pm$4.20}} &
        \textbf{60.60} {\scriptsize \textcolor{gray}{$\pm$3.80}} & 
        59.57 {\scriptsize \textcolor{gray}{$\pm$5.62}} &
        58.99 \\
        & \cellcolor{gray!30}{RELIEF} &
        \textbf{64.80} {\scriptsize \textcolor{gray}{$\pm$0.69}} &
        \underline{61.45} {\scriptsize \textcolor{gray}{$\pm$0.35}} &
        \textbf{55.03} {\scriptsize \textcolor{gray}{$\pm$0.39}} &
        \textbf{52.82} {\scriptsize \textcolor{gray}{$\pm$0.59}} &
        \textbf{79.75} {\scriptsize \textcolor{gray}{$\pm$1.28}} &
        \textbf{53.52} {\scriptsize \textcolor{gray}{$\pm$2.17}} &
        \underline{60.44} {\scriptsize \textcolor{gray}{$\pm$2.39}} &
        \textbf{62.97} {\scriptsize \textcolor{gray}{$\pm$1.19}} & 
        \textbf{61.35} \\
        
        \bottomrule
      \end{tabular}
  }
  \label{tab:graph_50shots}
  \vspace{-0.8em}
\end{table*}

\section{Related Work}

\subsection{Graph Prompt Tuning}
Graph prompt tuning research can be categorized based on whether the prompting methods are independent of pre-training strategies. For methods paired with specific pre-trained GNN models, GPPT ~\cite{GPPT} and GraphPrompt ~\cite{Gprompt} align pretext and downstream tasks using link prediction. All in One ~\cite{All_in_one} unifies tasks at graph level and incorporates learnable prompt graphs tuned by meta-learning. 
PSP~\cite{PSP} and Self-Pro~\cite{Self-Pro} further adapt prompting approaches to heterophilous graphs by introducing graph-level dual-view and asymmetric contrastive learning, respectively, as pretext tasks, preserving more graph structural information.

Although these methods have shown success, their reliance on specific pre-training strategies limits their applicability. In contrast, feature prompt tuning, has demonstrated comparable results across various pre-training strategies. Pioneering works like GPF, GPF-plus ~\cite{GPF}, and SUPT ~\cite{SUPT} introduce learnable feature prompts to every node. However, they potentially overemphasize prompts and disrupt the input space, which may degrade performance.

To the best of our knowledge, our work is the first to apply RL to graph feature prompt tuning, aiming to optimize a policy for inserting necessary and lightweight feature prompts to enhance the downstream tasks performance of pre-trained GNN models.

\subsection{RL for Graph Representation Learning}
The integration of RL into graph representation learning has advanced significantly. MAG-GNN ~\cite{MAG-GNN} uses DQN to select optimal subgraph subsets to achieve high expressivity. WSD ~\cite{WSD} employs weighted sampling for subgraph counting, with edge weights determined by DDPG. SUGAR ~\cite{SUGAR} preserves hierarchical graph properties by selecting subgraphs using an RL pooling mechanism based on Q-learning. GPA ~\cite{GPA} utilizes DQN to learn optimal annotation strategies for valuable nodes in active search, while GraphCBAL ~\cite{GraphCBAL} further employs A2C for a class-balanced active search strategy.

While these works typically apply basic RL algorithms to either select actions from finite sets or determine single real values, RELIEF employs RL in a hybrid action space, skillfully integrating policy generalization techniques to achieve intricate control, representing a more cohesive fusion of RL and graph learning.

\section{Experiments}

\subsection{Few-shot Graph Classification}
\paragraph{\textbf{GNN Architecture and Datasets.}} 
We use a 5-layer GIN~\cite{GIN} as the base architecture for the GNN model, which is pre-trained on chemistry datasets~\cite{chemistry} and prompt tuned on molecule properties prediction benchmark from MoleculeNet ~\cite{moleculenet} as downstream tasks. Descriptions of these datasets are provided in Appendix~\ref{sec:appendix_graph_data}.

\paragraph{\textbf{Pre-training Strategies.}}
To demonstrate the generality of our RELIEF, we pre-train the GIN model with four common strategies at both node and graph levels: Deep Graph Infomax (Infomax)~\cite{Infomax}, Attribute Masking (AttrMasking)~\cite{chemistry}, Context Prediction (ContextPred)~\cite{chemistry}, and Graph Contrastive Learning (GCL)~\cite{GCL}. Descriptions of these methods are available in Appendix~\ref{sec:appendix_graph_pre}.

\paragraph{\textbf{Baselines.}}
We compare RELIEF with the following tuning approaches, all of which involve adjusting the projection head. Besides, \textbf{Fine-Tuning} (FT) tunes the parameters of the pre-trained GNN model; \textbf{feature prompt tuning methods}, including GPF, GPF-plus~\cite{GPF}, SUPT\textsubscript{soft} and SUPT\textsubscript{hard}~\cite{SUPT}, tune the parameters of learnable feature prompts with the GNN model fixed; \textbf{other prompt-based method}, All in One~\cite{All_in_one}, uses a frozen GNN model specifically pre-trained by GCL, adapting learnable graph prompts, is included as it strictly aligns with our experimental setting.

\paragraph{\textbf{Training Scheme.}}
Since molecular predictions are multi-label binary classifications, we use binary cross-entropy loss as the reward function and evaluate by averaging the ROC-AUC of each label. We construct few-shot scenarios with 50 samples for prompt tuning over five random seeds. The number of sub-policies $l$ is set to 3, with penalty factors $\alpha_d$ selected from \{1e0, 1e1, 1e3, 1e5\} and $\alpha_c$ set to 0.1. The maximum prompt scale per step $z_\text{max}$ is selected from $\{0.1, 0.5, 1.0\}$. The projection head is chosen from $\{1, 2, 3\}$-layer MLPs and trained $q$ times per epoch, ranging from $\{1, 2, 3\}$. Detailed data splitting and hyper-parameter settings are provided in Appendix~\ref{sec:appendix_graph_exp_setings}. All the experiments were conducted on an NVIDIA A800 PCIe 80GB GPU.

\begin{table}[!tbp]
    \caption{PCR, APM and overall impact (OV.) of feature prompting methods averaged on all graph classification task.}
    \vspace{-3mm}
    \centering
    \fontsize{7.5pt}{7.5pt}\selectfont
        \begin{tabular}{ccccc}
        \toprule
        Tuning Strategy & PCR & APM ($10^{-2}$) & \textbf{OV.} ($10^{-2}$) & ROC-AUC  \\

        \midrule

        GPF & 1.00 & 7.15 & 7.15 & 62.08 \\
        GPF-plus & 1.00 & 6.69 & 6.69 & 62.38 \\
        SUPT\textsubscript{soft} & 1.00 & 6.46 & 6.46 & 62.39 \\
        SUPT\textsubscript{hard} & 0.65 & 6.10 & 3.97 & 62.68 \\
        \cellcolor{gray!30}{RELIEF} & 0.61 & 6.03 & 3.68 & 63.72 \\

        \bottomrule
             
        \end{tabular}
    \vspace{-5mm}
    \label{tab:metrics_result}
\end{table}

\paragraph{\textbf{Performance Results.}}
Based on the results in Table~\ref{tab:graph_50shots}, RELIEF achieves superior graph classification performance in few-shot settings, surpassing baselines in 28/32 tasks, with an average improvement of 1.64\% over fine-tuning and 1.04\% over the runner-up. Notably, RELIEF is the only method consistently surpassing fine-tuning across all tasks. Moreover, we observe that All in One yields better results by inserting prompt graphs with fewer nodes, suggesting that compact prompt structures may enhance prompt tuning.

\begin{figure*}[!t]
  \centering
  \includegraphics[width=0.9\textwidth]{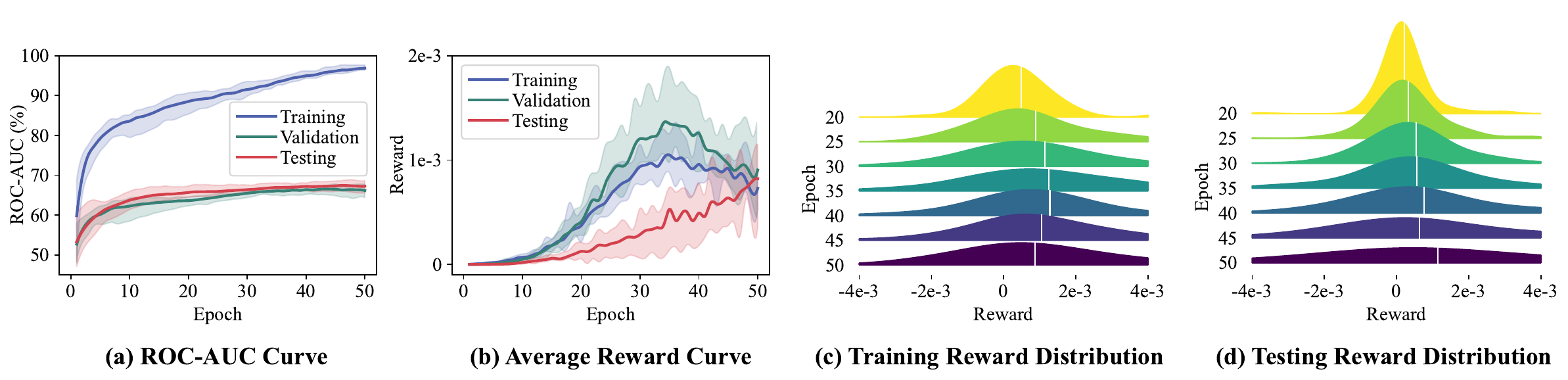}
  \vspace{-4mm}
  \caption{Tuning process of RELIEF on BACE. (a) and (b) present the ROC-AUC and average reward curves, respectively, for the training, validation, and testing sets across 5 random seeds. (c) and (d) illustrate the reward distributions for the training and testing sets as the tuning epochs progress, with white vertical lines indicating the average rewards for each distribution.}
  \label{fig:curves}
  \Description{Figure~\ref{fig:curves} Fully described in the text.}
  \vspace{-2mm}
\end{figure*}

We illustrate the tuning process of RELIEF on the BACE dataset pre-trained with Infomax in Figure~\ref{fig:curves}. The ROC-AUC curve demonstrates a monotonically increasing trend, indicating the stability of the training process. The reward curve and distributions show reward improvement as the epochs progress, confirming the effectiveness of RELIEF in enhancing downstream task performance.

Furthermore, we measure the impact of prompts on input graphs in terms of PCR and APM, and further multiply them to express the overall impact (OV.). Results are averaged across all tasks, as outlined in Table~\ref{tab:metrics_result}. RELIEF exhibits the smallest PCR, APM and overall impact compared to baselines. Interestingly, SUPT\textsubscript{hard} also prompts on specific nodes with a PCR of 0.65, as each basis prompt vector affects only the top-ranked nodes, where the top-rank ratio is a hyper-parameter. In contrast, RELIEF does not rely on priori information, such as the usage of basis vectors or a top-rank ratio, offering a highly flexible prompting policy based on the actors. Notably, the flexibility of graph manipulations is crucial for prompting performance~\cite{All_in_one, graph_prompt_survey}, contributing to RELIEF's superior results.

\subsection{Data Efficiency}
\label{sec:data_efficiency}
To evaluate the data efficiency of feature prompting methods, we incrementally add 5\% of the training samples until the average ROC-AUC over five random seeds surpasses full-shot fine-tuning. We report the minimum proportion of data required for each method on five datasets pre-trained with Infomax and ContextPred strategies.

\begin{figure}[!t]
    \centering
    \includegraphics[width=0.9\linewidth]{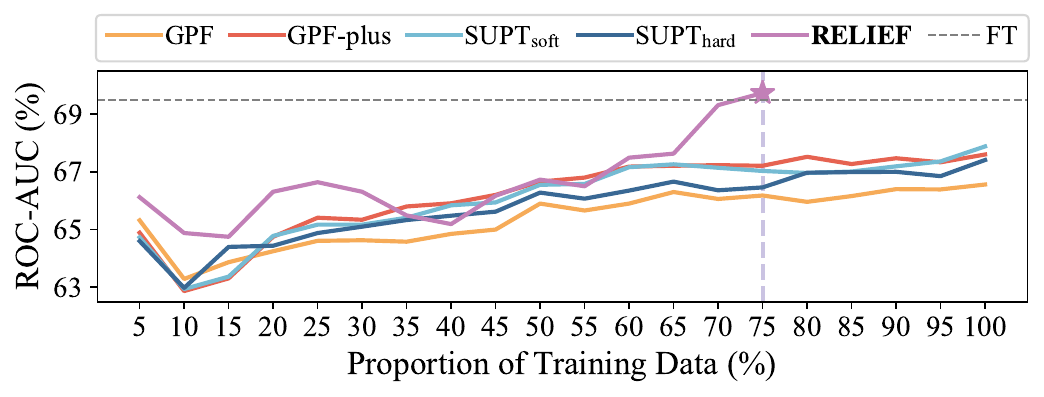}
    \vspace{-5mm}
    \caption{Performance of feature prompting methods with training data scaling up, compared to full-shot FT.}
    \label{fig:data_efficiency}
    \Description{Figure~\ref{fig:data_efficiency} fully described in the text.}
    \vspace{-4mm}
\end{figure}

Table~\ref{tab:data_efficiency} shows that RELIEF requires the least data for all tasks, while other baselines, more or less, fail to surpass full-shot fine-tuning even with the entire dataset. Figure~\ref{fig:data_efficiency} illustrates the performance trends on BBBP pre-trained with Infomax, which demonstrates that RELIEF achieves the most noticeable improvement as data scales up, whereas other methods fail to exceed fine-tuning.

We attribute RELIEF's superior data efficiency to its RL-based learning paradigm. Firstly, it reduces learning difficulty by optimizing one feature prompt at a time, allowing the agent to focus on a single state-to-action mapping instead of optimizing entire feature prompts at once. Secondly, the step-by-step insertion and evaluation of prompts expose the agent to various state patterns and loss values, acting as data augmentation in few-shot settings.

\begin{table}[!tbp]
    \caption{The minimum proportion (\%) of data required for surpassing the performance of full-shot FT. The minimum proportion are highlighted in bold. $\boldsymbol{\times}$ denotes cannot surpass. }
    \vspace{-3mm}
    \centering
    \fontsize{7.0pt}{7.0pt}\selectfont
    \begin{tabular}{ccccccc}
        \toprule
        & \makecell{Methods} & BBBP & Tox21 & SIDER & Cinltox & Bace \\

        \midrule
         
        \multirow{5}[2]{*}{\begin{sideways}Infomax\end{sideways}}
        & GPF & $\times$ & 60 & 60 & $\times$ & 55 \\ 
        & GPF-plus & $\times$ & 60 & 60 & 65 & 50 \\ 
        & SUPT\textsubscript{soft} & $\times$ & 65 & 60 & 70 & 45 \\ 
        & SUPT\textsubscript{hard} & $\times$ & 60 & 55 & 65 & 50 \\ 
        & \cellcolor{gray!30}{RELIEF} & \textbf{75} & \textbf{50} & \textbf{40} & \textbf{60} & \textbf{40} \\ 

        \addlinespace[1pt]
        \midrule[0.1pt]
        \addlinespace[1pt]

        \multirow{5}[2]{*}{\begin{sideways}ContPred\end{sideways}}
        & GPF & $\times$ & 55 & 55 & 70 & 95 \\ 
        & GPF-plus & $\times$ & \textbf{50} & 45 & 60 & 90 \\ 
        & SUPT\textsubscript{soft} & $\times$ & 65 & 50 & 65 & 85 \\ 
        & SUPT\textsubscript{hard} & $\times$ & 65 & 50 & 75 & 85 \\ 
        & \cellcolor{gray!30}{RELIEF} & \textbf{90} & \textbf{50} & \textbf{40} & \textbf{55} & \textbf{65} \\        
        \bottomrule
    \end{tabular}
    \label{tab:data_efficiency}
    \vspace{-3mm}
\end{table}

\subsection{Few-shot Node Classification}
\label{sec:node_exp}

\paragraph{\textbf{Extend to Node-Level Tasks.}} 
Given that feature prompting methods are originally designed for graph-level tasks~\cite{GPF}, adapting them to node-level tasks requires careful consideration. Fortunately, prior research has successfully unified various levels of tasks as subgraph-level tasks~\cite{Gprompt,All_in_one}. This unification is achieved by generating induced graphs for nodes or edges, which enables feature prompting methods to operate on $k$-hop subgraphs. These subgraphs act as proxies for the target nodes or edges during prompt tuning and downstream tasks evaluation.
 
\paragraph{\textbf{Datasets, Baselines and Pre-training Strategies.}}
We adopt Cora, Citeseer, Pubmed~\cite{Cora} and Amazon-Co-Buy (Computer and Photo)~\cite{Amazon} for node classification. To complement the node and graph-level pre-training strategies used in graph classification, here we focus on edge-level pre-training. Therefore, we include all previously mentioned feature prompting methods and additionally involve GPPT~\cite{GPPT} and GraphPrompt~\cite{Gprompt} as baselines, which are tailored to GNNs pre-trained at edge-level and capable of transferring to node classification. For pre-training strategies, GPPT uses masked edge prediction (MaskedEdge), while GraphPrompt employs contrastive learning based on edge connectivity (ContraEdge). Descriptions of the two strategies are presented in Appendix~\ref{sec:appendix_node}.

\paragraph{\textbf{Training Scheme.}}
We construct 10-shot node classification by randomly sampling 10 nodes per class, with additional 500 nodes each for validation and testing. Considering the average degree of different datasets, we set varying $k$ for generating k-hop subgraphs for each dataset. We employ a 2-layer GIN~\cite{GIN} as the fundamental GNN model. Cross-entropy loss is used as the reward function. Five rounds of experiments with different random seeds are conducted, and the performance is assessed via accuracy and macro F1-score. More experimental settings are available in Appendix~\ref{sec:appendix_node_exp}.

\paragraph{\textbf{Performance Results.}}
As illustrated in Table~\ref{tab:node_10shots}, RELIEF consistently achieves top-tier performance across majority of tasks, frequently ranking either first or as the runner-up, underscoring its applicability and effectiveness. On average, RELIEF outperforms fine-tuning by 2.05\% and 1.43\% on accuracy and macro f1-score, respectively. Additionally, RELIEF surpasses the second-best method, GPF-plus, by 0.89\% in accuracy. In contrast, GPPT and GraphPrompt exhibit inferior performance in most tasks. A plausible explanation is that their prediction results rely on class prototypes, which are challenging to accurately construct with limited labeled data.

\begin{table*}[htbp]
    \caption{Accuracy (\%) and Macro F1-score (\%) with respective standard deviation for node classification under 10-shot scenario with various two edge-level pre-training and various tuning strategies. The best results for each dataset and pre-training strategy are highlighted in bold, and the runner-up is underlined.}
    \vspace{-2mm}
    \resizebox{\linewidth}{!}{
        \centering
        \fontsize{6.2pt}{6.2pt}\selectfont
        \begin{tabular}{cc<{\hspace{3pt}}|>{\hspace{3pt}}c@{\hspace{3pt}}c|c@{\hspace{3pt}}c|c@{\hspace{3pt}}c|c@{\hspace{3pt}}c|c@{\hspace{3pt}}c}

        \toprule
        
        & \multirow{2}{*}{\makecell{Tuning \\ Strategy}} & \multicolumn{2}{c|}{Cora} & \multicolumn{2}{c|}{CiteSeer} & \multicolumn{2}{c|}{PubMed} & \multicolumn{2}{c|}{Computers} & \multicolumn{2}{c}{Photos} \\
        & & Accuracy & Macro F1 & Accuracy & Macro F1 & Accuracy & Macro F1 & Accuracy & Macro F1 & Accuracy & Macro F1 \\
    
        \midrule

        \multirow{7}[2]{*}{\begin{sideways}MaskedEdge\end{sideways}} &
        FT & 
        54.40 {\scriptsize \textcolor{gray}{$\pm$1.44}} &
        53.27 {\scriptsize \textcolor{gray}{$\pm$1.45}} &
        61.28 {\scriptsize \textcolor{gray}{$\pm$0.93}} &
        56.60 {\scriptsize \textcolor{gray}{$\pm$0.92}} &
        71.40 {\scriptsize \textcolor{gray}{$\pm$1.61}} &
        67.11 {\scriptsize \textcolor{gray}{$\pm$1.51}} &
        \underline{80.52} {\scriptsize \textcolor{gray}{$\pm$1.18}} &
        \underline{76.35} {\scriptsize \textcolor{gray}{$\pm$1.57}} &
        83.40 {\scriptsize \textcolor{gray}{$\pm$3.00}} &
        81.31 {\scriptsize \textcolor{gray}{$\pm$2.88}} \\
        & GPF &
        55.04 {\scriptsize \textcolor{gray}{$\pm$1.82}} &
        54.20 {\scriptsize \textcolor{gray}{$\pm$2.03}} &
        60.48 {\scriptsize \textcolor{gray}{$\pm$1.11}} &
        55.57 {\scriptsize \textcolor{gray}{$\pm$0.97}} &
        64.96 {\scriptsize \textcolor{gray}{$\pm$6.77}} &
        60.63 {\scriptsize \textcolor{gray}{$\pm$5.17}} &
        73.32 {\scriptsize \textcolor{gray}{$\pm$2.56}} &
        68.53 {\scriptsize \textcolor{gray}{$\pm$2.21}} &
        83.08 {\scriptsize \textcolor{gray}{$\pm$0.68}} &
        80.60 {\scriptsize \textcolor{gray}{$\pm$1.04}} \\
        & GPF-plus &
        56.36 {\scriptsize \textcolor{gray}{$\pm$1.46}} &
        55.84 {\scriptsize \textcolor{gray}{$\pm$1.33}} &
        \underline{61.36} {\scriptsize \textcolor{gray}{$\pm$0.70}} &
        56.50 {\scriptsize \textcolor{gray}{$\pm$0.44}} &
        \underline{72.32} {\scriptsize \textcolor{gray}{$\pm$4.17}} &
        \underline{67.52} {\scriptsize \textcolor{gray}{$\pm$3.51}} &
        \textbf{80.56} {\scriptsize \textcolor{gray}{$\pm$1.97}} &
        \textbf{76.69} {\scriptsize \textcolor{gray}{$\pm$1.87}} &
        85.08 {\scriptsize \textcolor{gray}{$\pm$1.22}} &
        82.26 {\scriptsize \textcolor{gray}{$\pm$0.94}} \\
        & SUPT\textsubscript{soft} &
        55.24 {\scriptsize \textcolor{gray}{$\pm$1.01}} &
        54.69 {\scriptsize \textcolor{gray}{$\pm$1.18}} &
        59.60 {\scriptsize \textcolor{gray}{$\pm$1.97}} &
        54.88 {\scriptsize \textcolor{gray}{$\pm$1.45}} &
        62.40 {\scriptsize \textcolor{gray}{$\pm$7.67}} &
        58.76 {\scriptsize \textcolor{gray}{$\pm$6.17}} &
        77.28 {\scriptsize \textcolor{gray}{$\pm$4.18}} &
        72.63 {\scriptsize \textcolor{gray}{$\pm$4.86}} &
        84.52 {\scriptsize \textcolor{gray}{$\pm$1.55}} &
        82.22 {\scriptsize \textcolor{gray}{$\pm$1.36}} \\
        & SUPT\textsubscript{hard} &
        \textbf{56.96} {\scriptsize \textcolor{gray}{$\pm$1.28}} &
        \underline{55.86} {\scriptsize \textcolor{gray}{$\pm$1.28}} &
        61.28 {\scriptsize \textcolor{gray}{$\pm$1.84}} &
        \underline{56.81} {\scriptsize \textcolor{gray}{$\pm$1.45}} &
        71.64 {\scriptsize \textcolor{gray}{$\pm$5.98}} &
        65.20 {\scriptsize \textcolor{gray}{$\pm$5.70}} &
        77.96 {\scriptsize \textcolor{gray}{$\pm$1.61}} &
        73.60 {\scriptsize \textcolor{gray}{$\pm$1.72}} &
        \textbf{85.24} {\scriptsize \textcolor{gray}{$\pm$0.81}} &
        \underline{82.92} {\scriptsize \textcolor{gray}{$\pm$0.56}} \\
        & GPPT &
        45.76 {\scriptsize \textcolor{gray}{$\pm$0.34}} &
        46.05 {\scriptsize \textcolor{gray}{$\pm$0.30}} &
        57.00 {\scriptsize \textcolor{gray}{$\pm$0.55}} &
        51.03 {\scriptsize \textcolor{gray}{$\pm$0.68}} &
        71.36 {\scriptsize \textcolor{gray}{$\pm$0.70}} &
        64.53 {\scriptsize \textcolor{gray}{$\pm$0.84}} &
        75.80 {\scriptsize \textcolor{gray}{$\pm$0.42}} &
        69.97 {\scriptsize \textcolor{gray}{$\pm$0.57}} &
        82.48 {\scriptsize \textcolor{gray}{$\pm$0.74}} &
        80.65 {\scriptsize \textcolor{gray}{$\pm$0.71}} \\
        & \cellcolor{gray!30}{RELIEF} &
        \underline{56.76} {\scriptsize \textcolor{gray}{$\pm$0.23}} &
        \textbf{55.94} {\scriptsize \textcolor{gray}{$\pm$0.25}} &
        \textbf{65.16} {\scriptsize \textcolor{gray}{$\pm$1.21}} &
        \textbf{58.55} {\scriptsize \textcolor{gray}{$\pm$1.12}} &
        \textbf{72.36} {\scriptsize \textcolor{gray}{$\pm$1.05}} &
        \underline{66.71} {\scriptsize \textcolor{gray}{$\pm$1.10}} &
        77.68 {\scriptsize \textcolor{gray}{$\pm$0.93}} &
        72.53 {\scriptsize \textcolor{gray}{$\pm$1.26}} &
        \underline{85.16} {\scriptsize \textcolor{gray}{$\pm$0.89}} &
        \textbf{84.07} {\scriptsize \textcolor{gray}{$\pm$0.69}} \\
        
        \addlinespace[1pt]
        \midrule[0.1pt]
        \addlinespace[2pt]
    
        \multirow{7}[2]{*}{\begin{sideways}ContraEdge\end{sideways}} &
        FT & 
        \underline{60.60} {\scriptsize \textcolor{gray}{$\pm$1.89}} &
        59.82 {\scriptsize \textcolor{gray}{$\pm$2.01}} &
        60.88 {\scriptsize \textcolor{gray}{$\pm$2.79}} &
        \underline{56.90} {\scriptsize \textcolor{gray}{$\pm$2.20}} &
        72.20 {\scriptsize \textcolor{gray}{$\pm$1.02}} &
        67.44 {\scriptsize \textcolor{gray}{$\pm$0.81}} &
        80.64 {\scriptsize \textcolor{gray}{$\pm$1.23}} &
        76.02 {\scriptsize \textcolor{gray}{$\pm$1.75}} &
        82.24 {\scriptsize \textcolor{gray}{$\pm$1.92}} &
        79.71 {\scriptsize \textcolor{gray}{$\pm$1.79}} \\
        & GPF &
        59.12 {\scriptsize \textcolor{gray}{$\pm$1.95}} &
        58.70 {\scriptsize \textcolor{gray}{$\pm$1.92}} &
        63.24 {\scriptsize \textcolor{gray}{$\pm$3.09}} &
        55.00 {\scriptsize \textcolor{gray}{$\pm$2.54}} &
        71.24 {\scriptsize \textcolor{gray}{$\pm$1.98}} &
        66.35 {\scriptsize \textcolor{gray}{$\pm$2.84}} &
        80.68 {\scriptsize \textcolor{gray}{$\pm$1.48}} &
        76.29 {\scriptsize \textcolor{gray}{$\pm$1.60}} &
        83.64 {\scriptsize \textcolor{gray}{$\pm$1.30}} &
        81.33 {\scriptsize \textcolor{gray}{$\pm$1.27}} \\
        & GPF-plus &
        \underline{60.60} {\scriptsize \textcolor{gray}{$\pm$1.08}} &
        \underline{60.63} {\scriptsize \textcolor{gray}{$\pm$0.87}} &
        62.44 {\scriptsize \textcolor{gray}{$\pm$1.86}} &
        56.81 {\scriptsize \textcolor{gray}{$\pm$0.55}} &
        \underline{73.64} {\scriptsize \textcolor{gray}{$\pm$3.73}} &
        \underline{69.95} {\scriptsize \textcolor{gray}{$\pm$3.67}} &
        80.68 {\scriptsize \textcolor{gray}{$\pm$0.65}} &
        \underline{76.93} {\scriptsize \textcolor{gray}{$\pm$0.92}} &
        86.20 {\scriptsize \textcolor{gray}{$\pm$1.15}} &
        83.57 {\scriptsize \textcolor{gray}{$\pm$1.19}} \\
        & SUPT\textsubscript{soft} &
        60.28 {\scriptsize \textcolor{gray}{$\pm$1.32}} &
        60.05 {\scriptsize \textcolor{gray}{$\pm$1.08}} &
        63.16 {\scriptsize \textcolor{gray}{$\pm$2.35}} &
        55.95 {\scriptsize \textcolor{gray}{$\pm$1.69}} &
        73.52 {\scriptsize \textcolor{gray}{$\pm$3.81}} &
        69.70 {\scriptsize \textcolor{gray}{$\pm$3.31}} &
        \underline{80.84} {\scriptsize \textcolor{gray}{$\pm$0.85}} &
        75.30 {\scriptsize \textcolor{gray}{$\pm$0.81}} &
        85.36 {\scriptsize \textcolor{gray}{$\pm$1.75}} &
        82.47 {\scriptsize \textcolor{gray}{$\pm$1.94}} \\
        & SUPT\textsubscript{hard} &
        58.92 {\scriptsize \textcolor{gray}{$\pm$0.86}} &
        58.75 {\scriptsize \textcolor{gray}{$\pm$0.82}} &
        \underline{64.32} {\scriptsize \textcolor{gray}{$\pm$2.38}} &
        55.96 {\scriptsize \textcolor{gray}{$\pm$2.20}} &
        73.56 {\scriptsize \textcolor{gray}{$\pm$1.97}} &
        69.48 {\scriptsize \textcolor{gray}{$\pm$2.18}} &
        80.52 {\scriptsize \textcolor{gray}{$\pm$0.95}} &
        76.69 {\scriptsize \textcolor{gray}{$\pm$0.87}} &
        \underline{86.28} {\scriptsize \textcolor{gray}{$\pm$0.55}} &
        \textbf{83.92} {\scriptsize \textcolor{gray}{$\pm$0.34}} \\
        & GPrompt &
        57.20 {\scriptsize \textcolor{gray}{$\pm$2.13}} &
        53.98 {\scriptsize \textcolor{gray}{$\pm$2.82}} &
        59.52 {\scriptsize \textcolor{gray}{$\pm$0.65}} &
        55.46 {\scriptsize \textcolor{gray}{$\pm$0.70}} &
        70.00 {\scriptsize \textcolor{gray}{$\pm$0.62}} &
        66.17 {\scriptsize \textcolor{gray}{$\pm$0.51}} &
        67.44 {\scriptsize \textcolor{gray}{$\pm$0.80}} &
        64.81 {\scriptsize \textcolor{gray}{$\pm$1.42}} &
        74.08 {\scriptsize \textcolor{gray}{$\pm$2.55}} &
        71.26 {\scriptsize \textcolor{gray}{$\pm$2.66}} \\
        & \cellcolor{gray!30}{RELIEF} &
        \textbf{61.20} {\scriptsize \textcolor{gray}{$\pm$1.20}} &
        \textbf{60.89} {\scriptsize \textcolor{gray}{$\pm$0.91}} &
        \textbf{66.28} {\scriptsize \textcolor{gray}{$\pm$0.27}} &
        \textbf{58.01} {\scriptsize \textcolor{gray}{$\pm$0.32}} &
        \textbf{74.96} {\scriptsize \textcolor{gray}{$\pm$1.72}} &
        \textbf{70.22} {\scriptsize \textcolor{gray}{$\pm$1.70}} &
        \textbf{82.04} {\scriptsize \textcolor{gray}{$\pm$0.83}} &
        \textbf{78.03} {\scriptsize \textcolor{gray}{$\pm$0.77}} &
        \textbf{86.48} {\scriptsize \textcolor{gray}{$\pm$0.90}} &
        \underline{83.88} {\scriptsize \textcolor{gray}{$\pm$0.62}} \\

        \bottomrule
        \end{tabular}
    }
    \label{tab:node_10shots}
    \vspace{-2mm}
\end{table*}

\subsection{Ablation Study}
\label{sec:appendix_ablation}
Considering the core of RELIEF lies in its two actors for incorporating prompts, we deactivated either one to create two variants. Specifically, ``random\_d'' randomly samples a valid node to prompt at each step, while ``random\_c'' generates a real-valued vector sampled from a random Gaussian distribution. Additionally, we include ``linear probing'', which discards the prompting components and adjusts only the projection head.

\begin{figure}[!t]
    \centering
    \includegraphics[width=0.9\linewidth]{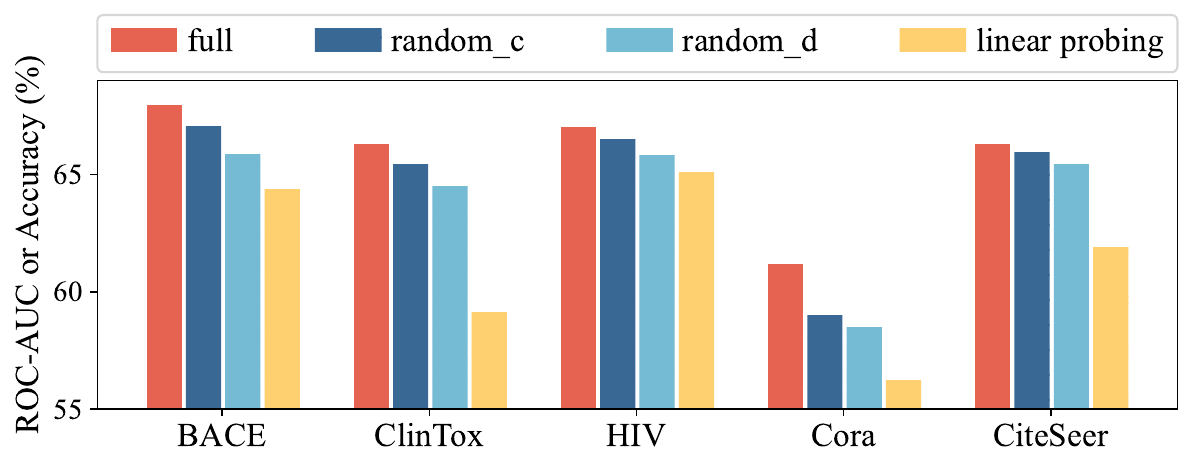}
    \vspace{-4mm}
    \caption{Effectiveness of discrete and continuous policies.}
    \label{fig:ablation}
    \Description{Figure~\ref{fig:ablation} Fully described in the text.}
    \vspace{-3mm}
\end{figure}

Figure~\ref{fig:ablation} shows the comparisons of RELIEF with its variants on three graph-level and two node-level tasks. We find that deactivating either actor results in performance drops due to the disruption of their coordination. Particularly, the discrete actor plays a crucial role in RELIEF. Intuitively, a random prompt context assigned to a target node can be rectified by prompting on the node again in later steps, given the premise of a powerful discrete actor for locating this node. However, with a random node selection policy, a suboptimal feature prompt is difficult to mitigate by prompting on arbitrary nodes, and is further exacerbated by the message-passing mechanism, which spreads chaotic prompts throughout the graph. In addition, "linear probing", which only trains the projection head, exhibits significantly inferior performance, clearly demonstrating the contribution of the inserted feature prompts. 

\begin{figure}[!tb]
    \centering
    \includegraphics[width=0.9\linewidth]{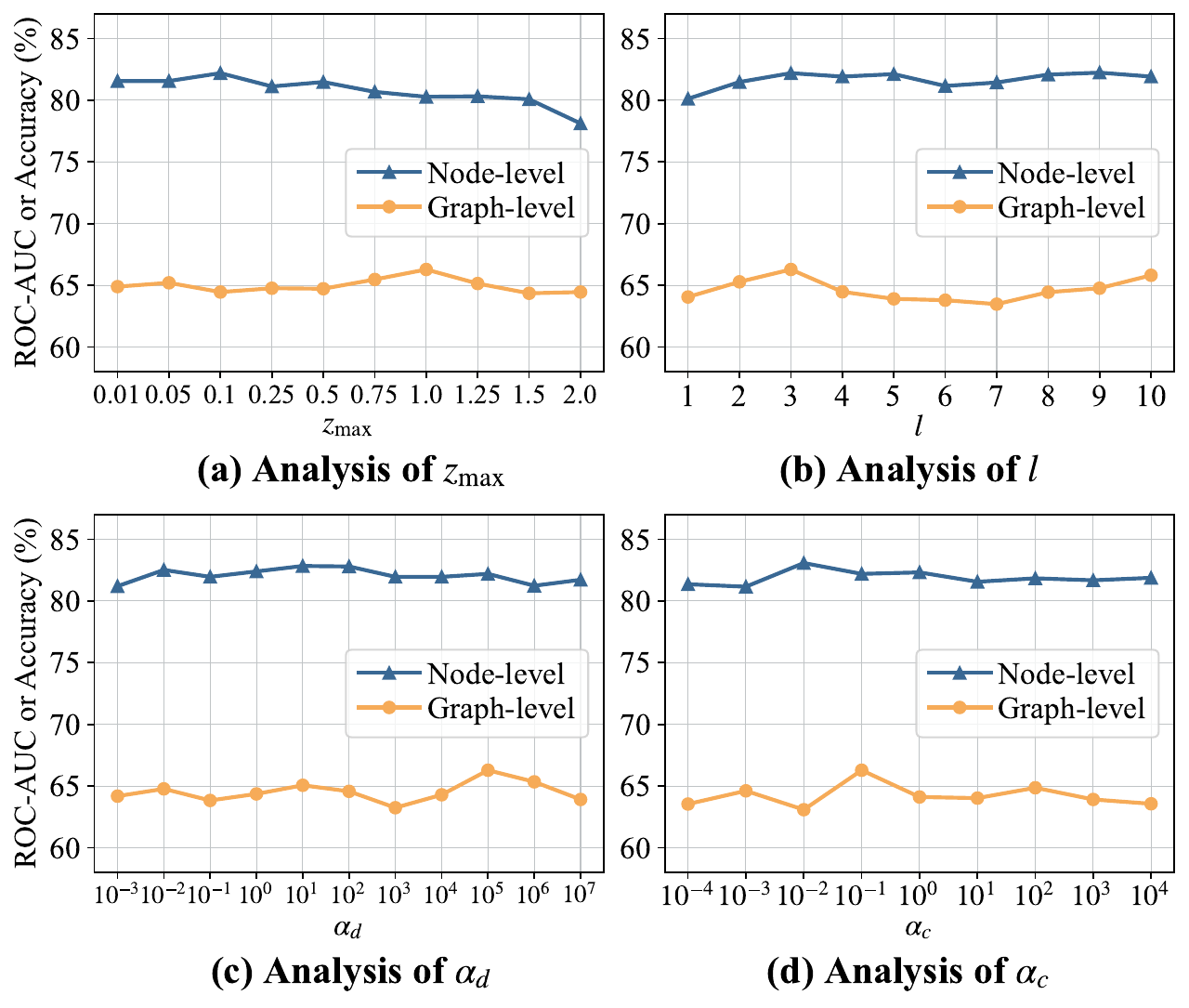}
    \vspace{-3mm}
    \caption{Parameter sensitivity analysis.}
    \label{fig:sensitivity}
    \Description{Figure~\ref{fig:sensitivity} fully described in the text.}
    \vspace{-5mm}
\end{figure}

\subsection{Parameter Analysis}
\label{sec:appendix_sensitivity}

We analyze the critical hyper-parameters in RELIEF, including the maximum feature prompt scale per step $z_\text{max}$, the number of sub-policies $l$, and the two penalty factors $\alpha_d$ and $\alpha_c$ related to policy generalization. Figure~\ref{fig:sensitivity} illustrates the ROC-AUC on ClinTox and accuracy on Computers, with each hyper-parameter varied over its legal range of values. The observations are as follows:

RELIEF performs consistently well across a broad range of hyper-parameter values, demonstrating its robustness. Specifically, when $z_\text{max}$ is too large, the excessive magnitude of prompts introduces unnecessary disturbance, leading to performance degradation. Conversely, setting $z_\text{max}$ too small weakens the impact of feature prompting, making it more similar to linear probing. By setting $l$ to 1, which discards the policy generalization technique, results in significant performance decline. While increasing $l$ may improve performance, it also requires maintaining more pairs of actors, leading to computational overhead. For $\alpha_d$ and $\alpha_c$, excessively large values may result in overly punishment, making the policies too conservative and hindering performance improvements.

Due to space constraints, we relocate the scalability analysis of our RELIEF to Appendix~\ref{sec:appendix_scalability}.

\subsection{Why RELIEF works?}
Despite of RELIEF's superior performance in graph and node-level tasks, it is essential to explain why RELIEF, utilizing an RL framework, fulfills our motivation -- achieving {\itshape powerful} downstream performance by adding {\itshape necessary} and {\itshape lightweight} prompts.

\textbf{\itshape Powerful.} 
RL demonstrates significant efficacy in solving combinatorial optimization problems~\cite{rl_combinatorial,MAG-GNN}. With a goal-oriented reward function and adequate exploration within stable state and reward dynamics, the RELIEF agent identifies an optimal prompting policy to maximize performance gains.

\textbf{\itshape Necessary.} The agent makes decisions based on state patterns and learned policies, displaying specific node selection preferences when faced with similar prompted features. Suboptimal node selection at any step leads to a cumulative reward decrease, which is captured by the critic and rectified through policy updates, thereby avoiding unnecessary prompts.

\textbf{\itshape Lightweight.} RELIEF tends to preserve pre-trained graph knowledge rather than overwhelm it. Larger-scale prompts cause the GNN model to generate unfamiliar representations, disrupting coordination between representations and the projection head, potentially leading to negative rewards.
These punishments are applied to the actor, urging it to provide prompts that avoid excessive disturbance to the initial input through policy updates.

\section{Conclusion}
In this paper, inspired by the marginal effect of increasing prompt token length on performance improvement in LLMs, we propose RELIEF, an RL empowered graph feature prompt tuning method. RELIEF enhances the downstream performance of pre-trained GNN models by incorporating only necessary and lightweight feature prompts to original graphs. To tackle this, we employ an RL algorithm designed for hybrid action spaces and integrated policy generalization techniques.
Extensive experiments on graph and node-level tasks demonstrate RELIEF's effectiveness and data efficiency compared to fine-tuning and other prompt-based methods.

\begin{acks}
This work is supported by National Natural Science Foundation of China No. 62202172.
\end{acks}

\clearpage

\bibliographystyle{ACM-Reference-Format}
\balance
\bibliography{base}

\clearpage
\appendix

\section{Details of Graph-level Tasks}
\label{sec:appendix_graph_exp}

For datasets and pre-training strategies used in graph classification, we adhere to the training procedures in~\cite{chemistry}, with pre-trained GIN models available on their GitHub repository \footnote{\url{https://github.com/snap-stanford/pretrain-gnns/tree/master/chem/model_gin}}.

\subsection{Datasets}
\label{sec:appendix_graph_data}
Chemical datasets~\cite{chemistry} are used to pre-train GNN models, consisting of both unlabeled and labeled molecules to facilitate comprehensive model training. Specifically, 2 million unlabeled molecules from the ZINC15 database are utilized for node-level self-supervised pre-training. For graph-level multi-task supervised pre-training, a preprocessed subset of the ChEMBL database including 456k labeled molecules is used, offering a diverse chemical space that enhances the model's ability to generalize across various biochemical tasks.

For downstream graph classification tasks, 8 binary classification datasets for molecular property prediction about biophysics and physiology~\cite{moleculenet} are employed. The statistics of these downstream datasets are summarized in Table~\ref{tab:graph_data}.

\begin{table}[h]
    \caption{Statistics of the downstream datasets used in our graph classification experiments.}
    \vspace{-2mm}
    \resizebox{\linewidth}{!}{
        \centering
        \fontsize{8pt}{8.5pt}\selectfont
        \begin{tabular}{ccc<{\hspace{5pt}}|>{\hspace{5pt}}ccc}
            \toprule
            Dataset & \# Molecules & \# Tasks & Dataset & \# Molecules & \# Tasks \\
            \midrule
            BBBP    & 2039 & 1   & ClinTox & 1478  & 2  \\
            Tox21   & 7831 & 12  & MUV     & 93087 & 17 \\
            ToxCast & 8575 & 617 & HIV     & 41127 & 1  \\
            SIDER   & 1427 & 27  & BACE    & 1513  & 1  \\
            \bottomrule
        \end{tabular}
    }
    \label{tab:graph_data}
    \vspace{-3mm}
\end{table}

\subsection{Pre-training Strategies}
\label{sec:appendix_graph_pre}
Four pre-training strategies are used to assess the universal applicability of prompting approaches.

Deep Graph Infomax (Infomax)~\cite{Infomax} aims to derive expressive representations for graphs or nodes by maximizing the mutual information between the overall graph-level representations and the substructure-level representations at various granularities. 

Attribute Masking (AttrMasking)~\cite{chemistry} involves obscuring certain node or edge attributes. GNN models are then tasked with predicting these masked attributes based on the surrounding structural information.

Context Prediction (ContextPred)~\cite{chemistry} uses subgraphs to predict their surrounding graph structures, which is designed to map nodes that appear in similar structural contexts to nearby embeddings, thereby capturing the context in which nodes are found.

Graph Contrastive Learning (GCL)~\cite{GCL} focuses on embedding augmented versions of the same anchor closely together (positive samples) while pushing the embeddings of different samples (negatives) apart.

The GNN models are first pre-trained using one of the above self-supervised methods and follow by the graph-level multi-task supervised pre-training.

\subsection{Experimental Settings}
\label{sec:appendix_graph_exp_setings}
We divide each dataset into training, validation, and testing sets with ratios of 80\%, 10\%, and 10\%, respectively. Molecules are divided based on their scaffolds, i.e., molecular graph substructures~\cite{scaffold}, and the clusters are recombined by prioritizing the most frequently occurring scaffolds in the training set. This produces validation and testing sets containing structurally different molecules~\cite{chemistry}, allowing us to evaluate the model's out-of-distribution generalization.

To create 50-shot scenarios, we randomly sampled 50 graphs from the training set. This procedure introduced randomness, differing from previous works~\cite{GPF,SUPT} that selected training data in descending order of scaffold occurrence. Given the limited budget of only 50 samples, prior methods resulted in training data with very few scaffold types, exacerbating overfitting. Consequently, the average ROC-AUC of fine-tuning across tasks under the 50-shot setting reported in previous works~\cite{GPF,SUPT} was 55.40$\pm$2.13\%, slightly above a random guess for binary classification tasks. In contrast, our randomness brings about a dataset with more molecular diversity and also prevents scaffold overlap between training, validation and test sets. As a result, the average ROC-AUC under our data splitting method is 60.92\% with a smaller standard deviation of 1.46\%.

We employ a 5-layer GIN with hidden dimensions of 300. The policy network and the projection head are independently updated using two Adam optimizers. We assess downstream task performance using the validation early stopping protocol, reporting the test ROC-AUC at the best validation epoch. Additionally, considering that 50-shot may be insufficient for some datasets to ensure stable performance at the onset of training, we determine the best validation epoch after 20 epochs. The complete hyper-parameters for graph classification are detailed listed in Table~\ref{tab:hyper_params}.

\section{Details of Node-level Tasks}
\label{sec:appendix_node}

\subsection{Datasets}
We adopt Cora, Citeseer, Pubmed~\cite{Cora} and Amazon-Co-Buy (Computer and Photo)~\cite{Amazon} for node classification. Cora, Citeseer, and Pubmed are citation networks, where each node represents a publication with a sparse bag-of-words feature vector, and edges signify citation links. Computer and Photo are segments of the Amazon co-purchase graph, with nodes representing products, edges indicating frequent co-purchases, node features encoded as bag-of-words from product reviews, and class labels derived from product categories. The statistics of these dataset are summarized in Table~\ref{tab:node_data}.

\begin{table}[h]
\caption{Statistics of the datasets used in our node classification experiments.}
    \vspace{-2mm}
    \centering
    \fontsize{7.5pt}{8pt}\selectfont
    \begin{tabular}{ccccc}
        \toprule
        Dataset & \# Nodes & \# Edges & \# Features & \# Classes \\
        \midrule
        Cora & 2708 & 5429 & 1433 & 7 \\
        CiteSeer & 3327 & 9104 & 3703 & 6 \\
        PubMed & 19717 & 88648 & 500 & 3 \\
        Computers & 13752 & 491722 & 767 & 10 \\
        Photo & 7650 & 238162 & 745 & 8 \\
        \bottomrule
    \end{tabular}
    \vspace{-3mm}
\label{tab:node_data}
\end{table}

\subsection{Pre-training Strategies}
To complement the node and graph-level pre-training strategies used in graph classification experiments, we focus on edge-level pre-training strategies to demonstrate the generality of RELIEF. Therefore, we include all previously mentioned feature prompting methods and additionally involve GPPT~\cite{GPPT} and GraphPrompt~\cite{Gprompt} as baselines, which are tailored to GNN models pre-trained at edge-level and capable of transferring to node classification.

\begin{table}[!htb]
    \caption{Hyper-parameter settings for graph and node level tasks.}
    \vspace{-2mm}
    \resizebox{\linewidth}{!}{
        \centering
        \fontsize{6.5pt}{7pt}\selectfont
        \begin{tabular}{l|cc}
             \toprule
             Hyper-parameter & graph-level & node-level \\
             \midrule
             \multicolumn{3}{c}{{\itshape General Hyper-parameters}} \\
             \addlinespace[2pt]
             \# Training epochs & \{50, 100\} & \{50, 100, 300\}  \\
             Graph loader batch size & \multicolumn{2}{c}{\{8, 16, 32, 64\}} \\
             \# GIN layers & 5 & 2 \\
             GIN hidden dimension & 300 & 128 \\
             Dropout ratio & \multicolumn{2}{c}{0.5} \\
             Actors learning rate (lr) & \multicolumn{2}{c}{5e-4} \\
             \# Sub-policies $l$ & \multicolumn{2}{c}{3} \\
             Discrete actors penalty $\alpha_d$ & \multicolumn{2}{c}{\{1e0, 1e1, 1e3, 1e5\}} \\
             Continuous actors penalty $\alpha_c$ & \multicolumn{2}{c}{0.1} \\
             Critic lr & \multicolumn{2}{c}{5e-4} \\
             Maximum prompt scale $z_\text{max}$ & \{0.1, 0.5, 1.0\} & \{0.05, 0.1, 0.5\} \\
             Policy network weight decay & \multicolumn{2}{c}{1e-5} \\
             Policy network lr linear decay & \multicolumn{2}{c}{\{True, False\}} \\
             Projection head lr & \multicolumn{2}{c}{\{5e-4, 1e-3, 1.5e-3\}} \\
             \# Projection head layers & \multicolumn{2}{c}{\{1, 2, 3\}} \\
             \# Projection head updates $q$ & \multicolumn{2}{c}{\{1, 2, 3\}} \\
             Projection head weight decay & 0 & 5e-4 \\
             Projection head lr linear decay & \multicolumn{2}{c}{\{True, False\}} \\
    
             \addlinespace[4pt]
             \multicolumn{3}{c}{{\itshape RL Hyper-parameters}} \\
             \addlinespace[2pt]
             Discount factor $\gamma$ & \multicolumn{2}{c}{0.99} \\
             GAE $\lambda$ & \multicolumn{2}{c}{0.95} \\
             \# PPO epochs & \multicolumn{2}{c}{10} \\
             PPO mini-batch size & \multicolumn{2}{c}{\{32, 64, 128, 256, 512\}} \\
             PPO clip range & \multicolumn{2}{c}{0.2} \\
             Return normalization & \multicolumn{2}{c}{True} \\
             Entropy bonus coefficient $\beta$ & \multicolumn{2}{c}{0.01} \\
             Critic loss coefficient & \multicolumn{2}{c}{0.5} \\
             \bottomrule
        \end{tabular}
    }
    \label{tab:hyper_params}
    \vspace{-3mm}
\end{table}

Formally, GPPT uses masked edge prediction (MaskedEdge) as pretext task, which is optimized by a binary cross-entropy loss $\mathcal{L}(\cdot)$ formulated as:
\begin{displaymath}
    \min_{\theta} \sum_{v_i,v_j} \mathcal{L} \left(f_\theta(v_i)^\top f_\theta(v_j), \mathbf{A}_{ij} \right)
\end{displaymath}
where $\mathbf{A}_{ij}$ represents if node $vi$ and $vj$ are connected given graph $\mathcal{G}$ with the adjacency matrix $\mathbf{A}$. Moreover, GraphPrompt employs contrastive learning using positive and negative node pairs determined based on edge connectivity (ContraEdge). The pre-training loss is defined as follows:
\begin{displaymath}
\begin{split}
    \min_{\theta} - \frac{1}{\left| \mathcal{V} \right|} & \sum_{v \in \mathcal{V}} \frac{1}{\left| \mathcal{V}^+ \right|} \cdot \\
    & \sum_{v^+ \in \mathcal{N}(v)} \log \frac{\exp (h_v^\top h_{v^+} / \tau)}{\exp (h_v^\top h_{v^+} / \tau) + \sum_{v^- \in \mathcal{V}^-} \exp (h_v^\top h_{v^-} / \tau)}
\end{split}
\end{displaymath}
where $h_v$, $h_{v^+}$ and $h_{v^-}$ are representations of node $v$, $v^+$ and $v^-$ produced by $f_\theta(\mathcal{G})$. $\mathcal{V}^+$ is the positive node set containing the neighbors of node $v$, while $\mathcal{V}^-$ is the negative node set constructed by random sampling. $\tau$ is the temperature hyper-parameter.

\subsection{Experimental Settings}
\label{sec:appendix_node_exp}
We construct 10-shot node classification experiments. Considering the average degree of different datasets, we set varying $k$ for generating k-hop subgraphs, specifically 4, 3, 2, 2, 2 for Cora, Citeseer, Pubmed, Computers and Photo, respectively, to ensure an appropriate number of nodes for induced subgraphs.

We employ a 2-layer GIN~\cite{GIN}, which is widely utilized in previous works~\cite{Cora,Amazon,GPPT}, with a hidden dimension of 128. While more advanced subgraph GNN techniques might yield better performance, as noted in \cite{GPF}, we opt to use the basic GIN for processing prompted subgraphs to ensure a fair comparison with GPPT and GraphPrompt, as these two methods do not incorporate subgraph-related designs. To achieve more compact node representations, we use Singular Value Decomposition (SVD) to reduce the initial features of each dataset to 100 dimensions. The policy network and the projection head are independently updated using two Adam optimizers. Other hyper-parameters are listed in Table~\ref{tab:hyper_params}. The RL hyper-parameter settings are partly referenced from \cite{PPO_hyper}, aligning with its established terminology.

\section{Scalability Analysis}
\label{sec:appendix_scalability}

We analyze the feasibility of scaling RELIEF to larger graphs from three perspectives.

\paragraph{\textbf{Performance.}} While larger graphs introduce additional challenges, they also provide more interactive transitions for updating the RL policy. We evaluate the average ROC-AUC improvement of RELIEF over fine-tuning across various datasets based on~\ref{tab:graph_50shots}. It can be observed datasets with more transitions yield larger performance gains. For example, BACE and MUV, with 1698 and 1243 transitions, respectively, achieve AUC improvements of 6.5 and 2.5, whereas Tox21 and ToxCast, with 806 and 805 transitions, show more modest gains of 0.2 and 1.0. This suggests that RELIEF can scale effectively with larger graphs.

\paragraph{\textbf{Memory.}} The only learnable modules in RELIEF are MLPs in actors and the critic. Given a node representation with a hidden size of $d$, the discrete actor's MLP maps tensors from $d$ to 1 representing the action probability, while the continuous actor's MLP maps from $d$ to $d$ for the prompt content. Thus, the number of parameters in the actors remains constant, independent of graph size. For the critic, we can further apply pooling operations to reduce the state representation size, which significantly decreases the input size for the critic’s MLP.

\paragraph{\textbf{Runtime.}} Table~\ref{tab:runtime} presents a comparison of average training time (in seconds) for 50 epochs between RELIEF and other prompt-based methods. While RL inevitably introduces additional complexity, the runtime remains manageable. Moreover, RELIEF's data efficiency reduces the need for extensive training data, potentially further shortening the overall training time.

\begin{table}[h]
\caption{Training time comparisons.}
    \centering
    \begin{tabular}{c|ccccc}
    \toprule
       Method & BBBP & Tox21 & ToxCast & ClinTox & BACE \\
    \midrule
       GPF & 110.1 & 112.0 & 108.7 & 109.3 & 108.8 \\
       GPF-plus & 113.2 & 116.9 & 112.6 & 110.4 & 111.5 \\
       SUPT-soft & 100.2 & 96.8 & 94.5 & 101.7 & 101.5 \\
       SUPT-hard & 100.3 & 99.5 & 94.8 & 100.1 & 99.3 \\
       All in one & 276.3 & 283.9 & 285.0 & 271.2 & 274.8 \\
       \textbf{RELIEF} & 225.6 & 220.8 & 188.5 & 297.2 & 284.3 \\
    \bottomrule
    \end{tabular}
    \label{tab:runtime}
\end{table}

\end{document}